\pdfoutput=1
\documentclass[sigconf,nonacm=true]{acmart}
\usepackage{multirow}
\usepackage{float,verbatim,enumitem}

\AtBeginDocument{%
  \providecommand\BibTeX{{%
    \normalfont B\kern-0.5em{\scshape i\kern-0.25em b}\kern-0.8em\TeX}}}

\setcopyright{acmcopyright}
\copyrightyear{2021}
\acmYear{2021}
\acmDOI{}

\acmConference[ACM'22]{ACM Conference}{2022}{USA}
\acmBooktitle{ACM '22: ACM Conference, 2022, USA}
\acmPrice{}
\acmISBN{}

\begin{document}


\title{Temporal Graph Network Embedding with Causal Anonymous Walks Representations}



\author{Ilya Makarov}
\email{iamakarov@hse.ru}
\orcid{0000-0002-3308-8825}
\affiliation{%
  \institution{HSE University}
  \city{Moscow}
  \country{Russia}
}
\affiliation{%
  \institution{University of Ljubljana}
  \city{Ljubljana}
  \country{Slovenia}
}

\author{Andrey Savchenko}
\email{avsavchenko@hse.ru}
\orcid{0000-0001-6196-0564}
\author{Arseny Korovko}
\author{Leonid Sherstyuk}
\affiliation{%
  \institution{HSE University}
  \city{Moscow}
  \country{Russia}
}

\author{Nikita Severin}
\affiliation{%
  \institution{Moscow Institute of Physics and Technology}
  \city{Moscow}
  \country{Russia}}
\email{severin.nn@phystech.edu}

\author{Aleksandr Mikheev}
\email{AGMikheev@sberbank.ru}
\affiliation{%
  \institution{HSE University}
  \city{Moscow}
  \country{Russia}
}
\affiliation{%
  \institution{Sber AI Lab}
  \city{Moscow}
  \country{Russia}
}

\author{Dmitrii Babaev}
\email{dmitri.babaev@gmail.com}
\orcid{0000-0002-3379-4308}
\affiliation{%
  \institution{Sber AI Lab}
  \city{Moscow}
  \country{Russia}
}



\begin{abstract}
 Many tasks in graph machine learning, such as link prediction and node classification, are typically solved by using representation learning, in which each node or edge in the network is encoded via an embedding. Though there exists a lot of network embeddings for static graphs, the task becomes much more complicated when the dynamic (i.e. temporal) network is analyzed. In this paper, we propose a novel approach for dynamic network representation learning based on Temporal Graph Network by using a highly custom message generating function by extracting Causal Anonymous Walks. For evaluation, we provide a benchmark pipeline for the evaluation of temporal network embeddings. This work provides the first comprehensive comparison framework for temporal network representation learning in every available setting for graph machine learning problems involving node classification and link prediction. The proposed model outperforms state-of-the-art baseline models. The work also justifies the difference between them based on evaluation in various transductive/inductive edge/node classification tasks. In addition, we show the applicability and superior performance of our model in the real-world downstream graph machine learning task provided by one of the top European banks, involving credit scoring based on transaction data.
\end{abstract}


\maketitle
\section{Introduction}
It is crucial for banks to be able to predict possible future interactions between companies: knowing that one company will be a client of another one gives an opportunity to offer financial and other siervices. It is also important to have comprehensive and meaningful information about each client. If this knowledge is expressed as client embeddings, then the problem of their compactness and expressiveness emerges. Banks own large datasets of financial interactions between their clients, which can be used for training and testing models solving link prediction.

Graph structures that describe dependencies between data are nowadays widely used to improve effectiveness of machine learning models trained on streaming data. In order to use conventional machine learning frameworks, it is necessary to develop a vector representation of the graph (such as network embeddings) by combining attributes from nodes (labels, text, etc.) and edges (i.e. weights, labels, timing) and taking into account the dynamic graph structure appearing in real-world problems. 

\begin{figure}[!ht]
 \centering
 \includegraphics[width=0.9\columnwidth]{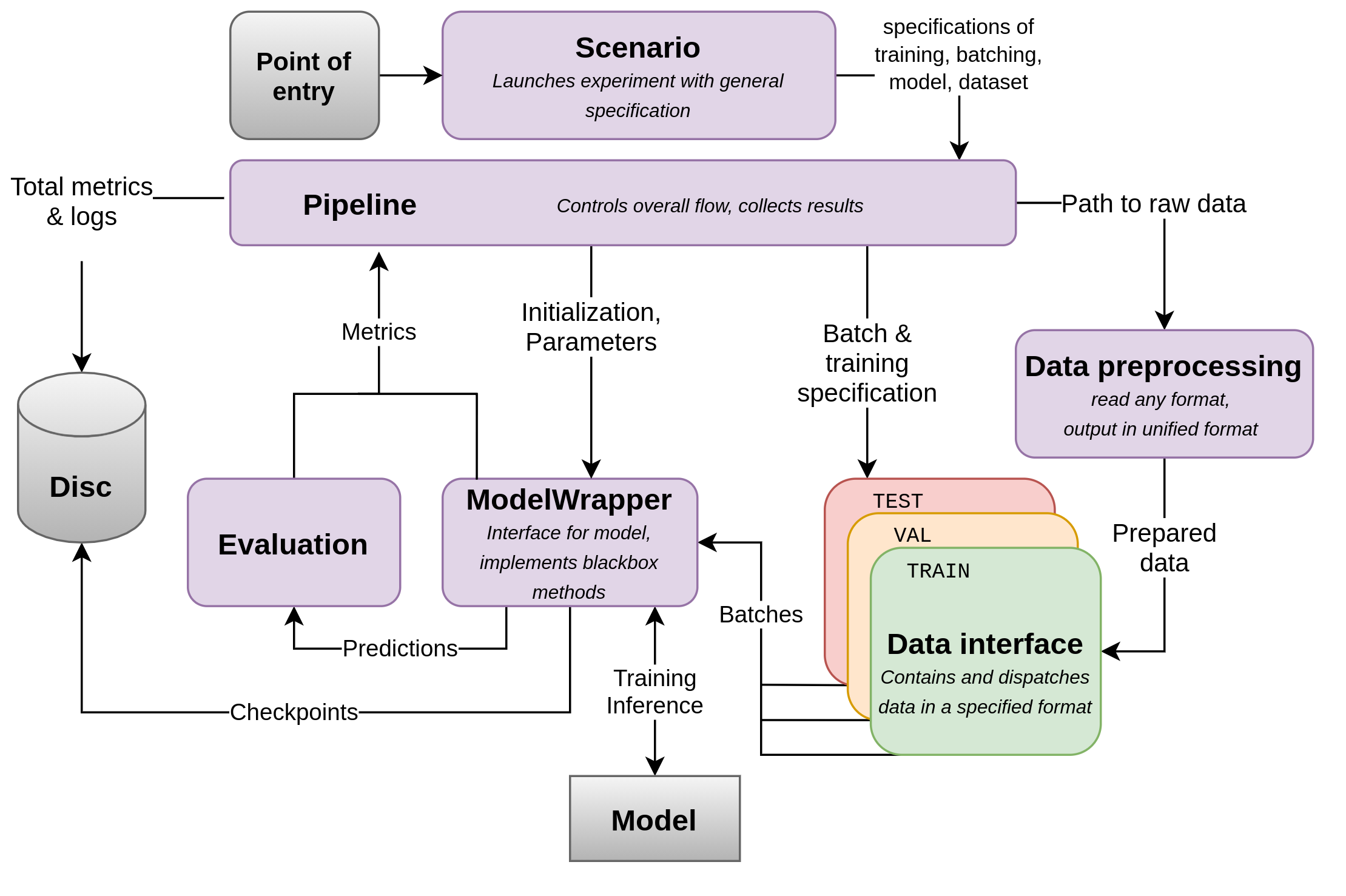}
 \caption{Diagram of the evaluation framework}
 \label{fig:pipeline}
\end{figure}

Existing graph embedding models have been actively studied in recent years in attempts to apply deep learning methods for network representation learning. Hundreds of graph embeddings models have been developed in computer vision, text processing, recommendation systems, and interdisciplinary research in biology and genetics  \cite{makarov2021survey}. All approaches are united by a common problem statement, which is to learn a model for the selected type of networks and important graph statistics. This will make it possible to apply standard machine learning frameworks and at the same time generalize attribute and structural information from the data.



Recently, modern machine learning methods have been proposed for processing networks and solving various prediction tasks on the local level. \cite{gcn, you2018graphrnn, gat, ding2018semi}. Examples include node classification and link prediction, for nodes both seen and unseen during the process of model training. In practice, they represent important problems to be solved on large dynamic network data, for e.g., transaction data between companies and bank clients can be used to predict future transactions; user search history on the Web can be used to generate contextual advertising instances; disease transmission data can be used to predict epidemics dynamics.

Although most of the previous works on graph representation learning mainly focus on static graphs (with a fixed set of nodes and edges), there are many real-world applications in which graphs evolve over time, like social networks or sales data. One particularly common sub-type of graphs used to represent such structures is a temporal graph. It is a graph in which each edge has a time index, indicating a moment in time when the interaction, represented by an edge, occurred. 

There are various problems when switching from static to dynamic networks \cite{graph_embed_issues}, among which computational complexity and variance of connectivity patterns over time. In addition, for practical applications, one needs to have models suitable for inference in inductive settings which enable proper prediction on the fly with rear overall model retraining for large network data \cite{dyngraph_survey}. 

In this work, we describe a novel network embedding which combines the best elements of the efficient Temporal Graph Network embedding (TGN)~\cite{tgn} with fast computed historic node embeddings and highly precise Causal Anonymous Walks (CAW)~\cite{caw} with attention over temporal edge dynamics. To properly evaluate the proposed model, we present an experimental framework for temporal network embedding evaluation in downstream graph machine learning tasks, allowing integration of various temporal network embeddings and different temporal network data under a unified evaluation framework shown in Fig.~\ref{fig:pipeline}.


Our main contributions with this work consist of the following:
\begin{enumerate}
\item Novel temporal network embedding model achieving state-of-the-art results in various temporal graph machine learning tasks;
\item Standardized temporal network embedding evaluation framework and comparison of state-of-the-art models under common training setting, providing new insights and clarification of real-world performance compared to reported in the original research articles.
\end{enumerate}

In addition, we prove the effectiveness of the proposed pipeline and its sub-modules via extensive ablation study and provide the industrial application of the proposed approach, involving the transaction data of a major European bank. We showed that feature enrichment of temporal attention over temporal edge random walks improves both quantitative and qualitative results in the real-world application of a banking graph machine learning task.

\section{Preliminaries \& Problem statement}

In order to proceed with the problem statement, we describe basic concepts used throughout the text following notations by \cite{dyngraph_survey}. We use $\mathcal{G}(\mathcal{V},\mathcal{E})$ to denote a static graph, where $\mathcal{V}$ is the set of its vertices, and $\mathcal{E}$ is the set of edges of the graph, and $\mathcal{A}$ is an adjacency matrix of that graph. 
Dynamic graph in general is a graph, which structure and node/edge attributes change over time. Generally, events may contain an updated state of the node, but in our experiments we consider all node features to be static, and represent them as $\mathcal{X} $ -- $|V| \times k $ matrix, where $k$ is the node feature dimensionality. 

In this work, we will use \textit{dynamic graph} and \textit{temporal graph} as interchangeable terms. We outline two possible kinds of dynamic graph representations below. There are two standard views on temporal network representation as a data structure: 
\begin{itemize}
\item {Discrete-time dynamic graph (DTDG)} 
or snapshot-like graph is a sequence of snapshots from a dynamic graph, sampled at regularly-spaced times. Formally, we define DTDG as a set \{$\mathcal{G}^1$, $\mathcal{G}^2$,..., $\mathcal{G}^T$, \}, where $\mathcal{G}^t = \{\mathcal{V}^t, \mathcal{E}^t \}$ is the graph at moment $t$, $\mathcal{V}^t$ and $\mathcal{E}^t$ are sets of nodes and edges in $\mathcal{G}^t$, respectively. 
\item {Continuous-time dynamic graph (CTDG)} 
or transaction-/stream-like graph is denoted by pair $(\mathcal{G},\mathcal{O})$, where G is a static graph representing an initial state of dynamic graph at time $t_0$, and $\mathcal{O}$ is event stream of events/transaction, denoted by tuple $(event\_coordinates, event\_data, timestamp)$, where $event\_coordinates$ are ordered pairs of nodes, between which the event has occurred, and $event\_data$ is any additional data on the event. 
\end{itemize}

Event stream may be represented as concatenation of  index matrix containing transactions vectors $e_{ij}(t)=(v_i,v_j, t)$ with source and target node IDs, timestamp, and temporal edges' features.

We will refer to both CTDG and DTDG as dynamic graphs, although we will focus more on CTDGs as a natural representation of transactions in banking networks, which appear in non-uniform timestamps and represent a real-world streaming structured data, rather than discretized snapshot representation DTDG.

Finally, in our study, we consider such temporal graph machine learning problems as \textit{node classification} and \textit{link prediction}, both in transductive and inductive settings:

\begin{itemize}
\item \textit{Transductive edge prediction} evaluates whether a transaction between two priorly known nodes occurred at a given time; 
\item \textit{Inductive edge prediction} predicts a transaction between known and unknown nodes at a given time;
\item \textit{Transductive node classification} determines the dynamic label of a priorly known node;
\item \textit{Inductive node classification} determines the dynamic label of a priorly unknown node.
\end{itemize}


All the mentioned above problems on the chosen datasets can be formulated as binary classification (with effortless extension to a multi-class case), which is evaluated via \textit{AUC-ROC} (area under curve for receiver operating characteristic) and \textit{AP} (average precision) quality metrics measuring performance for the classification at various error threshold settings. 


\section{Related work}
In this section, we overview state-of-the-art methods of constructing network embeddings for static and dynamic networks using taxonomies suggested in~\cite{makarov2021survey} and~\cite{barros2021survey}. We focus on dynamic network embedding as an evolutionary process of graph formation.

\subsection{Static graph embedding methods}
When constructing network embeddings via solving the optimization problem, researchers usually focus on three main concepts: matrix factorization, node sequence methods and methods based on deep learning.
We consider the snapshot method, in which the current snapshot of a temporal network is taken and missing links are predicted based on the available graph information.

Factorization techniques can be applied to different graph representations and optimized for different objectives, such as direct decomposition of the graph adjacency matrix~\cite{kruskal1978multidimensional,deerwester1990indexing, martinez2001pca} or approximating proximity matrix~\cite{roweis2000nonlinear, tenenbaum2000global}. Despite factorizations being widely used in recommendation systems, these models have high computational complexity and are difficult to extend for inductive learning. 

Inspired by word2vec~\cite{word2vec}, sequence-based methods aim to preserve local node neighborhoods based on random walks. The two most prominent examples of models in this class are DeepWalk~\cite{deepwalk} and Node2vec~\cite{node2vec}. Anonymous graphs walks have been proposed in~\cite{ivanov2018anonymous}. However, their adaptations to temporal random walks have limited applications, since they require retraining after adding new edges.

Recently, advances in geometric deep learning led to the creation of graph neural networks combining the best out of both fully connected and convolutional neural network architectures. Most of them use the method of neighborhood information aggregation from graph convolution network (GCN)~\cite{gcn} and extend it with classical deep learning architectures such as recurrent neural network (RNN)~\cite{you2018graphrnn}, attention mechanism~\cite{gat}, generative adversarial network (GAN)~\cite{ding2018semi}, and graph transformers~\cite{yun2019graph}.

Recent studies show that a combination of deep learning models with semi-supervised techniques gives state-of-the-art results in terms of scalability, speed and quality in downstream tasks~\cite{makarov2021survey}. However, static models are limited by the necessity to retrain the model with each significant change of graph structure.

\subsection{Dynamic graph embedding methods}
Methods for dynamic graphs are often extensions of those
for static ones, with an additional focus on the temporal dimension and update schemes~\cite{egcn}. All these techniques can be categorized according to which model of graph evolution representation is chosen: Continuous-Time Dynamic Graphs (CTDG) or Discrete-Time Dynamic Graphs (DTDG).

\subsubsection{DTDG-focused methods} Most of the early work on dynamic graph learning focuses exclusively on discrete-time dynamic graphs. Such models encode snapshots individually to create an array of embeddings or aggregate the snapshots in order to use a static method on them~\cite{Sharan2008}. All DTDG models can be divided into several categories, according to the approach they use for dealing with the temporal aspect of a dynamic graph.

\textit{Single-snapshot models}. Static models are used on the graph snapshots to make predictions for the next one (\cite{dne, to-gae, dyngem18}). Another way of implementing this methodology, called TI-GCN (Time Interval Graph Convolutional Networks) via residual architectures was proposed in~\cite{hisano2018semi, ti-gcn}. Besides single snapshots, these works use information from networks formation~\cite{hisano2018semi} represented by edge differences of several snapshots.

\textit{Multi-snapshot models}. In~\cite{tempnode2vec, dyn-vgae}, authors learn structural information of each snapshot by separate models. Authors of~\cite{tempnode2vec} compute individual sets of random walks for each snapshot in Node2Vec fashion and learn final node embeddings jointly, while in~\cite{dyn-vgae} autoencoders for each snapshot were trained in a consistent way to preserve similarity between consequent graph updates.

\textit{RNN-based models}. In contrast to previous methods, models in this category aim to capture sequential temporal dependencies mostly by feeding the output node embeddings or graph structure of each snapshot into RNN. Thus, GCN is combined with long short-term memory (LSTM) in~\cite{cd-gcn, gclstm, gcrn} or gated recurrent units (GRU) in~\cite{t-gcn, dcrnn}. Following these ideas, authors of~\cite{tna, resgnn} add residual connections to propagate topological information between neighboring snapshots. Recently, some works~\cite{gcn-gan, dyngan} have proposed to use GANs in combination with RNNs. On the other hand, EvolveGCN~\cite{egcn} argues that directly modeling dynamics of the node representation will hamper the model performance on graphs with dynamic node sets. Instead of treating node features as the input to RNN, it feeds the weights of the GCN into the RNN.

\textit{Temporal graph attention}. Inspired by advances in natural language processing (NLP), models in this class leverage attention-mechanism to capture temporal information. In~\cite{a3t-gcn, lrgcn, dyhatr}, authors follow the RNN module by an attention mechanism to take into account the contribution of its hidden states or leverage self-attention mechanism without RNN stage~\cite{dysat}.

\textit{Convolutional models}. Although previous models can capture sequential dependencies, in practice, most of them use a limited number of historical snapshots. So, convolutions are used to propagate information between snapshots in~\cite{ijcai2018-505, temporalgat, tsn}.

Despite promising results, most of the models struggle from two disadvantages. First, methods lose the order of edge formation and reflect only partial information on network formation~\cite{ti-gcn}. Second, computing static representations on each snapshot is inefficient in terms of memory usage on large graphs~\cite{cui2021dygcn} and can not be used in practical applications.

\subsubsection{CTDG-focused methods}
Continuous-time dynamic graphs require different approaches as it becomes computationally difficult to work with the entirety of such graphs after each interaction~\cite{goel2019}. Below we provide a more general classification of CTDG-focused methods, comparing with DTDG-focused ones based on approaches used for learning evolution patterns.

\textit{Temporal random-walks models}. The approach implies including the time dependency directly in a sequence of nodes generated by random walks. Such methods create a corpus of walks over time (so-called "temporal random walks") with respect to the order of nodes/edges appearance in the graph. Based on this idea, authors of~\cite{ehna} leverage a custom attention mechanism to learn the importance between a node and its temporal random-walk-based neighbors. Using this work,~\cite{ctdne} proposes several methods to select the subsequent nodes connected to a starting one. A promising method for link prediction task was proposed by the authors of~\cite{caw} who have developed Causal Anonymous Walks (CAWs), constructed from temporal random walks. CAWs adopt a novel anonymization strategy that replaces node identities with the hitting counts of the nodes based on a set of sampled walks to keep the method inductive. The model outperforms previous methods in both inductive and transductive settings.


\textit{Local neighborhood models}. When interactions happen (node or edge adding and removal), models in this class update embeddings of the relevant nodes by aggregating information from their new neighborhoods. In~\cite{cui2021dygcn, tdgnn}, authors utilize a GCN-based aggregation scheme and propagate changes to higher-order neighbors of interacting nodes. To cope with information asymmetry, the authors of~\cite{chen2021hili} propose to determine the priority of the nodes that receive the latest interaction information. Models from~\cite{jodie, tgat} embed dynamic network for recommendation systems in similar ways. Several recent works~\cite{dyrep, knyazev2019learning, mdne} consider interactions between nodes as stochastic processes with probabilities depending on the network statistics.

\textit{Memory-based models}. The core idea of this class of temporal network embeddings lies in extending existing models by using special memory modules storing a history of interactions for each node. Methods vary from updating LSTM~\cite{dgnn} to using augmented matrices of interactions~\cite{tigecmn}. APAN (Asynchronous Propagation Attention Network)~\cite{apan} aims to store detailed information about k-hop neighborhood interactions of each node in so-called "mailboxes". The recently developed temporal graph network~\cite{tgn} proposes a flexible framework, which consists of several independent modules and generalizes other recent CTDGs-focused models such as~\cite{tgat, jodie, dyrep}. Combining the advantages of JODIE~\cite{jodie} and temporal graph attention (TGAT)~\cite{tgat}, TGN introduces the node-wise memory into the temporal aggregate phase of TGAT, showing state-of-the-art for industrial tasks results.

Because of the potential of CTDG-based models~\cite{Gao2021OnTE} and necessity to apply the model to transaction data, we focus on developing a model in this class, while keeping in mind efficient DTDG models~\cite{ti-gcn}. In what follows, we describe the idea of improving existing state-of-the-art CTDG models by properly fusing memory-based, neighborhood and interaction information under unified framework, thus combining multiple best practices of CTDG methods.

\section{Proposed Approach}

As stated above in our literature survey, the best-known model for temporal graph prediction tasks, namely, TGN~\cite{tgn}, relies primarily on propagating information ("message") through edges and generates node embeddings from occurring edges in a straightforward manner. At the same time, it was noticed that the CAW~\cite{caw} shows excellent performance in link prediction tasks by learning edge representations. Unfortunately, the latter model cannot be directly applied to the extraction of node embeddings, and, consecutively, downstream graph machine learning tasks, such as node classification. 

\begin{figure}[!ht]
\centering
\includegraphics[width=0.9\columnwidth]{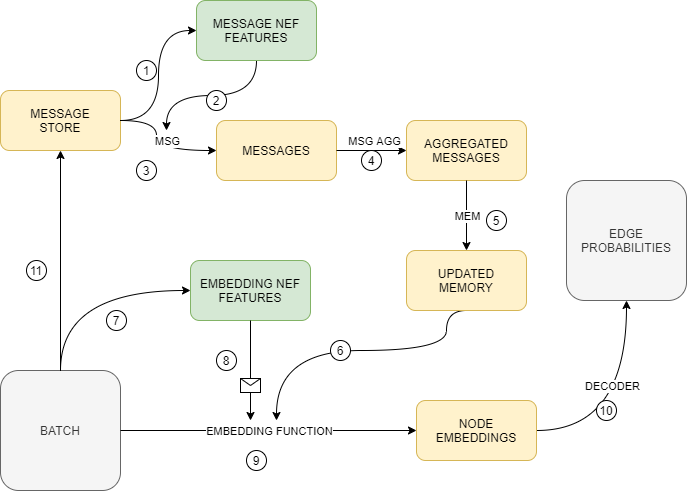}
\caption{Proposed model; numbers denote order of steps}
\label{fig:diagram}
\end{figure}

Hence, in this paper, we propose a novel network embedding taking the best out of the TGN framework and the CAW edge encodings (Fig.~\ref{fig:diagram}) to improve the quality of edge and node classification. Our contribution is to leverage the highly informative edge embedding generated in CAW. This will allow us to refine various functions of the TGN framework, including generating messages and embeddings, as well as aggregating memory. 

\subsection{Model details}
Below, we describe our model by listing and describing all modules in a consecutive fashion. In the following descriptions and equations, LSTM refers to the Long Short-Term Memory type RNN, GRU is the Gated Recurrent Unit, "attn" and "self-attn" is attention and self-attention, respectively. A walk on a graph, performed backwards in time on a temporal graph (time-inverse walk) with length $M$ is denoted by
\begin{multline}
W = ((w_0, t_0),(w_1, t_1), ..,(w_M, t_M)), t_0 > t_1 > ·· > t_M, \\ ({w_{m-1}, w_i}, t_m) \in \mathcal{E}, m \in \{1,2,...,M\}
\end{multline}
where $w_m \in \mathcal{V}$ is the node entered at step $m$, $t_m$ is the time of step $m$, $(w_m, t_m)$ is the $m$-th node-time pair. In future, we will refer to the elements of the $m$-th step as $W_m^{(1)} = w_m$ and $W_m^{(2)} = t_m$.

The $t$-timed $k$-hop neighborhood of $i$-th node is a set of nodes which can be reached from $i$ in $k$ steps if one walks on the edges existing prior to time $t$, without regard for their direction:
\begin{multline}
\eta_i^k = \{ w_j: \exists  W =  ((w_0, t_0),...,(w_j, t_j)), w_0 = i, \forall l: t_l < t,\\ ({w_{p-1}, w_p}, t_p) \in \mathcal{E} \vee ({w_p, w_{p-1}}, t_{p-1}) \in E, |W| \leq k+1 \}
\end{multline}

\subsubsection{Neighborhood Edge Feature (NEF) Generator)} \label{nef_gen}
This trainable module and its integration in the following modules are intended to improve the predictive capabilities of the original TGN framework, which is our core contribution. It generates highly informative feature representations of pairs of nodes for any given moment in time. For a pair of nodes $i$ and $j$, the output of this module as $NEF_{ij}(t)$ is computed as follows.

At first, we sample an equal number $K>1$ of the time-inverse walks for both nodes in order to capture information about the edge neighborhoods. All sampled walks have identical lengths (1-2 steps typically). A constant decay hyperparameter, which regulates how strongly the sampling process prioritizes more recent connections, is used during sampling of all walks.

The walk sets $S_{i}$ and $S_{j}$ are generated for edge $ij$ between nodes $i$ and $j$. $S_{i}$ consists of sampled time-inverse walks $W_k, k\in \{1,...,K\}$. Each edge for each walk is sampled with the probabilities proportional to $exp(\alpha \triangle t)$, where $\alpha$ is the decay parameter, and $\triangle t$ is the time difference between edges.

The next step makes the sampled walks anonymous in order to limit extracted information to the edge neighborhood alone. Each node from $S_{i}$ is replaced by a pair of vectors that encode a positional frequency of the node in each corresponding position of walks in $S_{i}$ and $S_{j}$. The encoding vector of positional frequencies relative to the walks of $i$ for node $w$ is defined as:
\begin{equation}
g(w, S_{i}) = \{|\{W | W \in S_{i}, w = W_m^{(1)}, m \in\{1,...,M\}\}|\}.
\end{equation}

This equation simply specifies that the node index $w$ is encoded as a vector, so that each its $m$-th component is equal to the number of times where $w$ is the $m$-th node of some walk in $S_i$.

The anonymization of walks~\cite{caw} is achieved by using the defined function to transform node indices in walks. Each position in any walk of $S_{i}$ containing $w$ is replaced by
\begin{equation}
I^{ij}(w) = \{ g(w, S_{i}), g(w, S_{j}) \}.
\end{equation}

Similarly, any value in walks of $S_{j}$ filled by $w$ is replaced with $I^{ji}(w)$. In the remaining part of this section we write $I = I^{ij}$, assuming the known orientation of the edge.

The remaining steps simply attempt to transform the ``anonymous walk" representation of the node pair to a more compressed and usable state. Each obtained representation of each position of each walk, i.e. pairs of $\{ g(w, S_{j}), g(w, S_{i}) \}$, are fed through separate instances of the same two-layered multi-layered perceptron (MLP) and sum-pooled:
\begin{equation}
f_1(I(w)) = \operatorname{MLP}(g(w, S_{i})) + \operatorname{MLP}(g(w, S_{j}))
\end{equation}

Then, this representation is concatenated with the time-difference encoding and node or edge features of the corresponding step:
\begin{equation}
h(I(w)) = \operatorname{concat}(f_1(I(w)), f_2(\triangle t), X)
\end{equation}
where $f_2(\triangle t)$ is time Fourier features, and $X$ is a concatenation of all relevant node and edge attributes of the corresponding walk step. Finally, each walk with encoded positions is passed to an RNN (usually, Bi-LSTM):
\begin{equation}
\operatorname{enc}(W) = \operatorname{Bi-LSTM}(\{ h(I(w_m)), m \in \{1,...,M\} \}),
\end{equation}
where $w_m$ being $m$-th position of walk $W$.
    
The encoded walks are aggregated across all walks for the node pair $ij$: 

\begin{equation}
NEF_{ij}(t) = \frac{1}{|S_i \cup S_j|} \sum_{W \in (S_i \cup S_j)}{ \operatorname{agg}(\operatorname{enc}(W)}),
\end{equation}
where $|S_i \cup S_j|$ is amount of walks in a set $S_i \cup S_j$, and $\operatorname{agg}$ is either self-attention module or identity for mean pooling aggregation.

\subsubsection{Message store}\label{msg_store}
In order to apply the gradient descent, a memory of a node should be updated after it is passed as a training instance. Let's say an event associated with the $i$-th node has occurred, e.g. an edge involving $i$ has been passed for training in the current batch. Then, all information about the batch transactions involving node $i$ will be recorded in the message store after the batch inference, replacing the existing records for $i$. This information is used and updated during the processing of the next batch involving $i$. 

\subsubsection{Message Generator}\label{msg_gen}
When the model processes a batch containing a node, all transaction information about edges associated with this node is pulled. For each transaction between $i$ and some other node $j$ at a time $t$, a message for a node $i$ is computed as a concatenation of the current memory vectors of the nodes, edge features, time-related features, and neighborhood edge features of the corresponding edge:
\begin{equation}
m_i(t) = concat(s_i(t^-), s_j(t^-), t-t^-, e_{ij}(t), NEF_{ij}(t)),
\end{equation}
where $ij$ is the index of a transaction between nodes $i$ and $j$, $s_i(t^-)$ is a memory state of node $i$ at time $t^-_i$ of last memory update for node $i$, and $NEF_{ij}(t)$ is a neighborhood edge feature vector for edge $ij$. The usage of NEF features here is our novel idea aimed to include more information about the type of the update message. The generated messages will update memory states of the batch nodes. Note that, aside from the basic and non-trainable concatenation, other choices for the message function, like MLPs are possible.
    
\subsubsection{Message Aggregator}\label{msg_agg}
To perform a memory update on a node at time $t$, its message representation is obtained by aggregating all currently stored messages timestamps $t_1 < t_2 < ... < t$ which are related to this node:
\begin{equation}
\overline{m}_i(t) = \operatorname{aggr}(m_i(t_1), m_i(t_2), ..., m_i(t)).
\end{equation}

Here aggregation function $aggr$ can be computed as a mean of generated messages. It is still possible to use the most recent message $m_i(t)$ as a value of $\overline{m}_i(t)$.
    
\subsubsection{Memory Updater}\label{mem_upd}
The message generator and aggregator let the model to encode useful transaction information as memory vectors. In particular, the memory state vector for node $i$ is updated at time $t$ by applying an RNN-type model to the concatenation of the received message and previous memory state:
\begin{equation}
s_i(t) = \operatorname{RNN}(\overline{m}_i(t), s_i(t^-))
\end{equation}
where $\operatorname{RNN}$ is either GRU or LSTM, and initial state $s(0)$ is initialized with a random vector.
    
\subsubsection{Embedding Generator}\label{emb_gen}
This module generates the node embedding based on memory states of the node and its k-hop neighborhood, features of the neighborhood, and NEF representations of "virtual" edges between the node and its direct neighbors. Resulting node embeddings can be viewed as autonomous node representations for classification. We propose, similarly to message generator, to add NEF features to node $i$ in order to let the model better discriminate neighbors on their relevance or type:
\begin{multline}
z_i(t) = \sum_{j \in \eta_i^1} h_0(s_i(t), s_j(t), e_{ij}, v_i(t), v_j(t), NEF_{ij}(t)) + \\ \sum_{j \in \eta_i^k, k > 1} h_1(s_i(t), s_j(t), e_{ij}, v_i(t), v_j(t)), 
\end{multline}
where $v_i(t)$ is a feature vector of node $i$ at time $t$, and $h_0$ and $h_1$ are the units of the neural network. It is typical to use MLPs as $h_0$ and $h_1$, but the following concatenation by self-attention is more preferable to capture complex dependencies involving NEFs.
    
\subsubsection{Embedding Decoder}\label{emd_dec}
This is the final module, which transforms node embeddings into prediction results for downstream tasks. In this paper, we always use MLP with 3 layers with only node embeddings as inputs, and sigmoid or softmax output layers. For example, the multi-class classification problem for node $i$ at time $t$ is solved using the following equation:
\begin{equation}
out_i(t) = \operatorname{Softmax}(\operatorname{MLP}(z_i(t))).
\end{equation}

Similarly, the edge prediction task is solved as follows:
\begin{equation}
out_{ij}(t) = \operatorname{Sigmoid}(\operatorname{MLP}(\operatorname{concat}(z_i(t) z_j(t)))).
\end{equation}

\subsection{Core idea and novelty of proposed model}

Let us provide a high-level description of the model (Fig.~\ref{fig:diagram}). It learns to generate a temporal embedding for each node and decode embeddings into inputs for each classification task. 
The model assigns a memory vector to each node and generates each node embedding by aggregating memory vectors and other relevant features in a neighborhood of the node. Node memory vectors describe relevant information about interactions involving the node. They are updated using specified node messages, which, in turn, encode information about the last transaction involving the node. This design allows employing gradient descent for training memory and message generating modules. 
A decoder MLP transforms a pair of node embeddings into the probability of a temporal edge existing between the nodes. Similarly, it can be trained to transform a single node embedding into probabilities of the node belonging to each of the existing classes for a node classification problem. 

There are two main modifications of the proposed model compared to the baseline TGN framework~\cite{tgn}. First, we change the message generating function, which provides the model with an additional way to differentiate the messages based on their relevance. The NEF features of an edge contain information about the walk correlation of the two nodes. As it can be used for very accurate link prediction~\cite{caw}, it may be also used to classify some messages as being irrelevant, and diminish their effect on the memory update. 

Second, while the original version of the embedding module~\cite{tgn} allows treating different k-hop neighbors of the node differently if using attention for aggregation, it might be beneficial to provide the NEF features of connections between the node and its closest neighbors. As a result, we again take into account the walk correlation between the node and its neighbor, so that the differences in neighborhood type can be more evident.

\begin{table*}[t]
\centering
\caption{Dscriptive statistics for the dataset. Left to right: whether the graph is bipartite, number of unique nodes (representing users and items) and edges, labels, average degree, number of edge updates per node as source/target, and setting for splitting the temporal network into batches for snapshot DTDG models}
\label{tab:data_desc}
\resizebox{\textwidth}{!}{
\begin{tabular}{lccccccc}\hline
\multicolumn{1}{l}{\textbf{}} & \multicolumn{1}{l}{\textit{Bipartite}} & \multicolumn{1}{l}{\textit{\begin{tabular}[c]{@{}l@{}}Unique  nodes\end{tabular}}} & \multicolumn{1}{l}{\textit{Unique edges}} & \multicolumn{1}{l}{\textit{Positive labels / Fraction}} & \multicolumn{1}{l}{\textit{Average degree}} & \textit{\begin{tabular}[c]{@{}c@{}}Average edges per n  ode\end{tabular}} & \textit{Batch setting}          \\ \hline
\textbf{Reddit}                & Yes                                    & 10000 / 984                                                                                      & 672447                                    & 366 / 0.0005                                            & 61.372                                      & 67.244 / 683.381                                                                               & 1 day per batch (30 days total) \\
\textbf{Wikipedia}             & Yes                                    & 8227 / 1000                                                                                      & 157474                                    & 217 / 0.0013                                            & 17.192                                      & 19.141 / 157.474                                                                               & 1 day per batch (30 days total) \\
\textbf{Enron}                 & No                                     & 185                                                                                              & 125236                                    & -                                                       & 29.059                                      & 675.778                                                                                        & 1\% of total edge count         \\
\textbf{UCI}                   & No                                     & 1900                                                                                             & 59836                                     & -                                                       & 106.405                                     & 13.043                                                                                         & 1\% of total edge count         \\
\textbf{Ethereum}              & No                                     & 231288                                                                                           & 300000                                    & -                                                       & 2.339083                                    & 2.59416                                                                                        & 1\% of total edge count\\\hline        
\end{tabular}
}
\end{table*}

\section{Evaluation framework}
In this section, we discuss our research methodology. This involves the evaluation framework, evaluation pipeline training settings, hyper-parameter choice and description of temporal networks used for the evaluation.

\subsection{Pipeline} The main contribution of the proposed framework (Fig.~\ref{fig:pipeline}) is an easy-to-use unified data processing toolkit for accurate evaluation of temporal network embeddings in downstream tasks under common training settings, which allows to remove contradictions in experimental results reported in research articles in the field. The source code of this tool is publicly available\footnote{\url{https://github.com/HSE-DynGraph-Research-team/DynGraphModelling}}. It is focused on transforming any graph into a universal format that is afterwards fed into the pipeline in either DTDG or CTDG format with following features:
\begin{itemize}
\item selection of precise batching options;
\item preparing data for both inductive and transductive settings (with support for bipartite data); 
\item interfacing with any kind of graph embedding models using an interface for model-framework communication, treating network embedding models as a black-box, which is then passed on to such basic methods as \textit{train\_model}, \textit{predict\_edge\_probabilities}, etc.
\end{itemize}

\subsection{Training settings}

\subsubsection{Negative sampling} It is common to design include negative edge sampling into evaluation of temporal graph prediction tasks. It is employed for balancing edges and non-connected pairs of nodes. We follow a standard setting, where the number of negative samples is equal to the number of edges. 

\subsubsection{Batch specification} The model is trained by passing edges (which includes negative samples) in batches sorted in chronological order. 

\subsubsection{Training with temporal data masking} We use 80\%-10\%-10\% train-val-test split with randomized training sets, supported via several data masking schemes. \textit{Node masking} hides nodes (as well as all connected edges) from training data. The masked nodes for transductive tasks are also removed from the validation and test data, while for inductive tasks masked nodes remain in validation and test data. \textit{Edge masking} removes a fixed percentage of random edges from the whole dataset. We use two options for both masking techniques: the percentage of masked information is 10\% and 70\%, representing dense and sparse training settings, respectively. In particular, we report the results for three combinations, in which at least one mask is large, namely 10\%-75\% node-edge masking, 75\%-10\% and 75\%-75\%.

\subsubsection{Runs and validation} Each model/setting/dataset combination was run 10 times with random seeds and node/edge masking. 


\subsection{Comparison with the state-of-the-art models}


In order to evaluate our model (Fig.~\ref{fig:diagram}), we compare it to the baseline TGN framework architectures (configured as described in~\cite{tgn}), DyRep (supporting long-term and short-term time scales of graph evolution and scaling to large graphs)~\cite{dyrep} and Jodie (working with bipartite graphs using node embeddings for predictive tasks with future time lag)~\cite{jodie}. Though these models can construct high-quality historic embeddings using information propagated from nodes, they do not take into account edge information and interaction patterns. Hence, we could demonstrate the advantages of our approach in identical training settings in comparison to other state-of-the-art temporal network embeddings. It is important to emphasize that the CAW model cannot be used here because it was created only for graph reconstruction tasks and cannot produce node/edge embeddings for the downstream tasks.



\subsection{Datasets}
In our study, we focus on well-known benchmark datasets for temporal networks. Their descriptive statistics are presented in Table \ref{tab:data_desc}, taking into account specifics of computing statistics for bipartite graphs of user-item interactions \cite{bipartite_stats}.

\paragraph{Labeled datasets}
Both \textit{Reddit} \cite{graphsage} and \textit{Wikipedia} \cite{dumpsWikimedia} datasets are bipartite graphs representing interactions between users (source) and web resources (target), like posting to a subreddit or editing a wiki page. Text content is encoded as edge features to provide context. Both datasets have a binary target, indicating whether a user was banned or not.

\paragraph{Non-labeled datasets}
The \textit{UCI} \cite{cortez2008using} dataset contains a communication history for the students' forum. The \textit{Enron} \cite{leskovec2008community} dataset is constructed over internal e-mail communication of Enron company employees. The \textit{Ethereum} dataset\footnote{\url{https://github.com/blockchain-etl/ethereum-etl}} contains a directed graph of Ethereum blockchain transactions, which were collected from larger public BigQuery dataset\footnote{\url{https://cloud.google.com/bigquery/public-data}}; we collected a dataset containing all transactions, which occurred during a 9-hour span. All of the non-labeled datasets are non-bipartite, with no additional features for edges or nodes.






    
    
    
    
    


\section{Ablation study}
In this Section an ablation study of the proposed architecture (Fig.~\ref{fig:diagram}) is provided to explore the impact of each submodule. 
Below we specify several additional modules added to interaction features processing pipeline for temporal network embedding via NEF generator,which outlined in Subsection~\ref{nef_gen}:
\begin{itemize}[align=left]\label{modifications}
\item[(Msg)]  NEF-Message concatenating NEF features and messages;
\item[(Emb)] NEF-Embed generates embeddings with NEF features;
\item[(RNN)] NEF-LSTM including Bi-LSTM into NEF generator, instead of mean pooling across walk elements.
\end{itemize}

Additionally, we consider alternating three important NEF-related hyper-parameters:
\begin{itemize}
\item number of generated random walk NEF-samples;
\item random walk depth;
\item positional dimension of random-walk-related embeddings. 
\end{itemize}

Table \ref{tab:ablation} contains AUC-ROC standard deviation and mean averaged 10 times. The task measured was inductive edge prediction with default hyperparameters. Rather large and representative Reddit dataset is used in ablation study. Below we provide performance metrics for seven different combinations of modules mentioned above, except "(RNN)", which is considered a standalone TGN with NEF features processing regularization. 
Here the resulting model significantly improves the performance of standalone modifications and TGN baseline. 

\begin{table}[t]
\centering
\caption{Ablation study on Reddit dataset for inductive edge prediction}\label{tab:ablation}
\begin{tabular}{lc}\hline
Enabled modules   & AUC-ROC \\ \hline
(Msg)+(Emb)+(RNN) & $0.894 \pm 0.034$             \\
(Emb)             & $0.893 \pm 0.035$             \\
(Msg)             & $0.888 \pm 0.042$             \\
(Emb)+(Msg)       & $0.878 \pm 0.051$             \\
(Msg)+(RNN)       & $0.876 \pm 0.047$             \\
(Emb)+(RNN)       & $0.870 \pm 0.055$             \\
TGN baseline      & $0.865 \pm 0.065$             \\
\hline
\end{tabular}
\end{table}


\section{Experimental results}
In this section, we compute mean AUC-ROC and AP for our model and compare it to three other state-of-the-art models (DyRep, Jodie, TGN). Tables~\ref{tab:t_e_auc}-Table \ref{tab:i_e_ap} show results for edge prediction task in transductive and inductive settings respectively. Best results are given in bold, while second best are underlined. "Node mask" and "Edge mask" columns specify the portion of nodes or edges used for model testing. 



\begin{table}[H]
\centering
\caption{Transductive edge prediction, AUC-ROC; best in bold, second-best underlined}
\label{tab:t_e_auc}
\resizebox{\columnwidth}{!}{
\begin{tabular}{lccccccc}\hline
                                                         & \multicolumn{1}{c}{\textit{\textbf{Node mask}}} & \multicolumn{1}{c}{\textit{\textbf{Edge mask}}} & \multicolumn{1}{c}{\textbf{DyRep}} & \multicolumn{1}{c}{\textbf{Jodie}} & \multicolumn{1}{c}{\textbf{TGN}} &  \multicolumn{1}{c}{\textbf{\begin{tabular}[c]{@{}c@{}}Our\end{tabular}}} \\ \hline
\multicolumn{1}{l}{\multirow{3}{*}{\textbf{Enron}}}     & \multicolumn{1}{c}{\textit{10\%}}              & \textit{75\%}                                  & \textbf{0.774}                      & 0.534                               & 0.638                                                                                                                          & \underline{0.759}                                                                                                \\ %
\multicolumn{1}{l}{}                                    & \multicolumn{1}{c}{\textit{75\%}}            & \textit{10\%}                                    & 0.647                               & 0.528                               & \underline{0.682}                                                                                                                    & \textbf{0.795}                                                                                                \\ %
\multicolumn{1}{l}{}                                    & \multicolumn{1}{c}{\textit{75\%}}            & \textit{75\%}                                  & 0.595                               & 0.537                               & \underline{0.673}                                                                                                                             & \textbf{0.745}                                                                                                \\ \hline
\multicolumn{1}{l}{\multirow{3}{*}{\textbf{Reddit}}}    & \multicolumn{1}{c}{\textit{10\%}}              & \textit{75\%}                                  & 0.964                               & 0.704                               & \underline{0.974}                                                                                                                    & \textbf{0.977}                                                                                                \\ 
\multicolumn{1}{l}{}                                    & \multicolumn{1}{c}{\textit{75\%}}            & \textit{10\%}                                    & 0.973                               & 0.825                               & \textbf{0.977}                                                                                                                  & \textbf{0.977}                                                                                                \\ 
\multicolumn{1}{l}{}                                    & \multicolumn{1}{c}{\textit{75\%}}            & \textit{75\%}                                  & 0.959                               & 0.72                                & \underline{0.976}                                                                                                                    & \textbf{0.977}                                                                                                \\ \hline
\multicolumn{1}{l}{\multirow{3}{*}{\textbf{UCI}}}       & \multicolumn{1}{c}{\textit{10\%}}              & \textit{75\%}                                  & 0.77                                & 0.496                               & \underline{0.804}                                                                                                                  & \textbf{0.808}                                                                                                \\ 
\multicolumn{1}{l}{}                                    & \multicolumn{1}{c}{\textit{75\%}}            & \textit{10\%}                                    & \underline{0.731}                               & 0.573                               & 0.709                                                                                                                    & \textbf{0.812}                                                                                                \\ 
\multicolumn{1}{l}{}                                    & \multicolumn{1}{c}{\textit{75\%}}            & \textit{75\%}                                  & \underline{0.781}                      & 0.489                               & 0.776                                                                                                                            & \textbf{0.808}                                                                                                \\ \hline
\multicolumn{1}{l}{\multirow{3}{*}{\textbf{Wikipedia}}} & \multicolumn{1}{c}{\textit{10\%}}              & \textit{75\%}                                  & 0.966                               & 0.728                               & \underline{0.968}                                                                                                                  & \textbf{0.969}                                                                                                \\ 
\multicolumn{1}{l}{}                                    & \multicolumn{1}{c}{\textit{75\%}}            & \textit{10\%}                                    & 0.964                               & 0.737                               & \underline{0.971}                                                                                                                   & \textbf{0.975}                                                                                                \\ 
\multicolumn{1}{l}{}                                    & \multicolumn{1}{c}{\textit{75\%}}            & \textit{75\%}                                  & 0.962                               & 0.719                               & \underline{0.967}                                                                                                                  & \textbf{0.97}                                                                                                 \\ \hline
\multicolumn{1}{l}{\multirow{3}{*}{\textbf{Ethereum}}} & \multicolumn{1}{c}{\textit{10\%}}              & \textit{75\%}                                  & 0.728                               & 0.914                               & \underline{0.932}                                                                                                                  & \textbf{0.933}                                                                                                \\ 
\multicolumn{1}{l}{}                                    & \multicolumn{1}{c}{\textit{75\%}}            & \textit{10\%}                                    & 0.868                               & 0.929                               & \underline{0.936}                                                                                                                   & \textbf{0.937}                                                                                                \\ 
\multicolumn{1}{l}{}                                    & \multicolumn{1}{c}{\textit{75\%}}            & \textit{75\%}                                  & 0.73                               & 0.919                               & \textbf{0.934}                                                                                                                  & \textbf{0.934}                                                                                                 \\ \hline
\end{tabular}
}
\end{table}

\begin{table}[H]
\centering
\caption{Transductive edge prediction, AP; best in bold, second-best underlined}
\label{tab:t_e_ap}
\resizebox{\columnwidth}{!}{
\begin{tabular}{lccccccc}\hline
                                                         & \multicolumn{1}{c}{\textit{\textbf{Node mask}}} & \multicolumn{1}{c}{\textit{\textbf{Edge mask}}} & \multicolumn{1}{c}{\textbf{DyRep}} & \multicolumn{1}{c}{\textbf{Jodie}} & \multicolumn{1}{c}{\textbf{TGN}} &  \multicolumn{1}{c}{\textbf{\begin{tabular}[c]{@{}c@{}}Our\end{tabular}}} \\ \hline
\multicolumn{1}{l|}{\multirow{3}{*}{\textbf{Enron}}}     & \multicolumn{1}{l|}{\textit{Lean}}              & \textit{Strict}                                  & \textbf{0.772}                      & 0.517                               & 0.651                             & \underline{0.762}                                                                                                \\ \cline{2-3}
\multicolumn{1}{l|}{}                                    & \multicolumn{1}{l|}{\textit{Strict}}            & \textit{Lean}                                    & 0.668                               & 0.527                               & \underline{0.685}                             & \textbf{0.787}                                                                                                \\ \cline{2-3}
\multicolumn{1}{l|}{}                                    & \multicolumn{1}{l|}{\textit{Strict}}            & \textit{Strict}                                  & 0.6                                 & 0.524                               & \underline{0.699}                    & \textbf{0.753}                                                                                                \\ \hline
\multicolumn{1}{l|}{\multirow{3}{*}{\textbf{Reddit}}}    & \multicolumn{1}{l|}{\textit{Lean}}              & \textit{Strict}                                  & 0.965                               & 0.688                               & \underline{0.973}                            & \textbf{0.977}                                                                                                \\ \cline{2-3}
\multicolumn{1}{l|}{}                                    & \multicolumn{1}{l|}{\textit{Strict}}            & \textit{Lean}                                    & 0.974                               & 0.832                               & \textbf{0.977}                    & \textbf{0.977}                                                                                                \\ \cline{2-3}
\multicolumn{1}{l|}{}                                    & \multicolumn{1}{l|}{\textit{Strict}}            & \textit{Strict}                                  & 0.96                                & 0.699                               & \textbf{0.977}              & \textbf{0.977}                                                                                                \\ \hline
\multicolumn{1}{l|}{\multirow{3}{*}{\textbf{UCI}}}       & \multicolumn{1}{l|}{\textit{Lean}}              & \textit{Strict}                                  & 0.804                               & 0.503                               & \textbf{0.827}               & \underline{0.826}                                                                                                \\ \cline{2-3}
\multicolumn{1}{l|}{}                                    & \multicolumn{1}{l|}{\textit{Strict}}            & \textit{Lean}                                    & 0.756                               & 0.546                               & \underline{0.75}                           & \textbf{0.824}                                                                                                \\ \cline{2-3}
\multicolumn{1}{l|}{}                                    & \multicolumn{1}{l|}{\textit{Strict}}            & \textit{Strict}                                  & \underline{0.812}                      & 0.492                               & 0.809                & \textbf{0.827}                                                                                                \\ \hline
\multicolumn{1}{l|}{\multirow{3}{*}{\textbf{Wikipedia}}} & \multicolumn{1}{l|}{\textit{Lean}}              & \textit{Strict}                                  & 0.969                               & 0.708                               & \underline{0.971}   & \textbf{0.972}                                                                                                \\ \cline{2-3}
\multicolumn{1}{l|}{}                                    & \multicolumn{1}{l|}{\textit{Strict}}            & \textit{Lean}                                    & 0.966                               & 0.711                               & \underline{0.974}       & \textbf{0.976}                                                                                                \\ \cline{2-3}
\multicolumn{1}{l|}{}                                    & \multicolumn{1}{l|}{\textit{Strict}}            & \textit{Strict}                                  & 0.965                               & 0.695                               & \underline{0.969}       & \textbf{0.973}                                                                                                \\ \hline
\end{tabular}
}
\end{table}

\begin{table}[H]
\centering
\caption{Inductive edge prediction, AUC-ROC; best in bold, second-best underlined}
\label{tab:i_e_auc}
\resizebox{\columnwidth}{!}{
\begin{tabular}{lccccccc}\hline
                                                         & \multicolumn{1}{c}{\textit{\textbf{Node mask}}} & \multicolumn{1}{c}{\textit{\textbf{Edge mask}}} & \multicolumn{1}{c}{\textbf{DyRep}} & \multicolumn{1}{c}{\textbf{Jodie}} & \multicolumn{1}{c}{\textbf{TGN}} &  \multicolumn{1}{c}{\textbf{\begin{tabular}[c]{@{}c@{}}Our \end{tabular}}} \\ \hline
\multicolumn{1}{l}{\multirow{3}{*}{\textbf{Enron}}}     & \multicolumn{1}{c}{\textit{10\%}}              & \textit{75\%}                                  & \underline{0.613}                               & 0.453                               & 0.591                                                                                                                    & \textbf{0.677}                                                                                                \\ 
\multicolumn{1}{l}{}                                    & \multicolumn{1}{c}{\textit{75\%}}            & \textit{10\%}                                    & 0.672                               & 0.508                               & \underline{0.705}                                                                                                                    & \textbf{0.792}                                                                                                \\ 
\multicolumn{1}{l}{}                                    & \multicolumn{1}{c}{\textit{75\%}}            & \textit{75\%}                                  & 0.666                               & 0.573                               & \underline{0.702}                                                                                                                    & \textbf{0.766}                                                                                                \\ \hline
\multicolumn{1}{l}{\multirow{3}{*}{\textbf{Reddit}}}    & \multicolumn{1}{c}{\textit{10\%}}              & \textit{75\%}                                  & 0.832                               & 0.588                               & \underline{0.823}                                                                                                                   & \textbf{0.836}                                                                                                \\ 
\multicolumn{1}{l}{}                                    & \multicolumn{1}{c}{\textit{75\%}}            & \textit{10\%}                                    & 0.91                                & 0.294                               & \underline{0.917}                                                                                                           & \textbf{0.924}                                                                                                \\ 
\multicolumn{1}{l}{}                                    & \multicolumn{1}{c}{\textit{75\%}}            & \textit{75\%}                                  & 0.88                                & 0.298                               & \underline{0.907}                                                                                                                   & \textbf{0.917}                                                                                                \\ \hline
\multicolumn{1}{l}{\multirow{3}{*}{\textbf{UCI}}}       & \multicolumn{1}{c}{\textit{10\%}}              & \textit{75\%}                                  & \underline{0.602}                      & 0.576                               & 0.599                                                                                                                    & \textbf{0.642}                                                                                                \\ 
\multicolumn{1}{l}{}                                    & \multicolumn{1}{c}{\textit{75\%}}            & \textit{10\%}                                    & 0.723                               & 0.431                               & \underline{0.725}                                                                                                                  & \textbf{0.733}                                                                                                \\ 
\multicolumn{1}{l}{}                                    & \multicolumn{1}{c}{\textit{75\%}}            & \textit{75\%}                                  & 0.718                               & 0.423                               & \underline{0.735}                                                                                                                   & \textbf{0.74}                                                                                                 \\ \hline
\multicolumn{1}{l}{\multirow{3}{*}{\textbf{Wikipedia}}} & \multicolumn{1}{c}{\textit{10\%}}              & \textit{75\%}                                  & \underline{0.95}                       & 0.584                               & 0.945                                                                                                                            & \textbf{0.953}                                                                                                \\ 
\multicolumn{1}{l}{}                                    & \multicolumn{1}{c}{\textit{75\%}}            & \textit{10\%}                                    & \underline{0.902}                               & 0.462                               & 0.899                                                                                                                   & \textbf{0.918}                                                                                                \\ 
\multicolumn{1}{l}{}                                    & \multicolumn{1}{c}{\textit{75\%}}            & \textit{75\%}                                  & 0.888                      & 0.47                                & \underline{0.894}                                                                                                                            & \textbf{0.911}                                    \\ \hline
\multicolumn{1}{l}{\multirow{3}{*}{\textbf{Ethereum}}} & \multicolumn{1}{c}{\textit{10\%}}              & \textit{75\%}                                  & 0.519                               & 0.655                               & \underline{0.677}                                                                                                                  & \textbf{0.679}                                                                                                \\ 
\multicolumn{1}{l}{}                                    & \multicolumn{1}{c}{\textit{75\%}}            & \textit{10\%}                                    & 0.483                               & 0.629                               & \underline{0.656}                                                                                                                   & \textbf{0.667}                                                                                                \\ 
\multicolumn{1}{l}{}                                    & \multicolumn{1}{c}{\textit{75\%}}            & \textit{75\%}                                  & 0.494                               & 0.616                               & \textbf{0.654}                                                                                                                  & \textbf{0.654}                                                                                                 \\ \hline
\end{tabular}
}
\end{table}

\begin{table}[H]
\centering
\caption{Inductive edge prediction results, AP; best in bold, second-best underlined}
\label{tab:i_e_ap}
\resizebox{\columnwidth}{!}{
\begin{tabular}{lccccccc}\hline
                                                         & \multicolumn{1}{c}{\textit{\textbf{Node mask}}} & \multicolumn{1}{c}{\textit{\textbf{Edge mask}}} & \multicolumn{1}{c}{\textbf{DyRep}} & \multicolumn{1}{c}{\textbf{Jodie}} & \multicolumn{1}{c}{\textbf{TGN}} &  \multicolumn{1}{c}{\textbf{\begin{tabular}[c]{@{}c@{}}Our \end{tabular}}} \\ \hline

\multicolumn{1}{l}{\multirow{3}{*}{\textbf{Enron}}}     & \multicolumn{1}{c}{\textit{10\%}}              & \textit{75\%}                                  & \underline{0.649}                      & 0.527                               & 0.627                            & \textbf{0.689}                                                                                                \\ 
\multicolumn{1}{l}{}                                    & \multicolumn{1}{c}{\textit{75\%}}            & \textit{10\%}                                    & 0.668                               & 0.524                               & \underline{0.712}                            & \textbf{0.785}                                                                                                \\ 
\multicolumn{1}{l}{}                                    & \multicolumn{1}{c}{\textit{75\%}}            & \textit{75\%}                                  & 0.678                               & 0.566                               & \underline{0.71}                          & \textbf{0.765}                                                                                                \\ \hline
\multicolumn{1}{l}{\multirow{3}{*}{\textbf{Reddit}}}    & \multicolumn{1}{c}{\textit{10\%}}              & \textit{75\%}                                  & \underline{0.858}                      & 0.582                               & 0.851          & \textbf{0.86}                                                                                                 \\ 
\multicolumn{1}{l}{}                                    & \multicolumn{1}{c}{\textit{75\%}}            & \textit{10\%}                                    & 0.924                               & 0.393                               & \underline{0.931}      & \textbf{0.936}                                                                                                \\ 
\multicolumn{1}{l}{}                                    & \multicolumn{1}{c}{\textit{75\%}}            & \textit{75\%}                                  & 0.899                               & 0.391                               & \underline{0.917}      & \textbf{0.928}                                                                                                \\ \hline
\multicolumn{1}{l}{\multirow{3}{*}{\textbf{UCI}}}       & \multicolumn{1}{c}{\textit{10\%}}              & \textit{75\%}                                  & \underline{0.664}                      & 0.576                               & 0.663                & \textbf{0.69}                                                                                                 \\ 
\multicolumn{1}{l}{}                                    & \multicolumn{1}{c}{\textit{75\%}}            & \textit{10\%}                                    & 0.762                               & 0.46                                & \underline{0.763}       & \textbf{0.771}                                                                                                \\ 
\multicolumn{1}{l}{}                                    & \multicolumn{1}{c}{\textit{75\%}}            & \textit{75\%}                                  & 0.756                               & 0.461                               & \textbf{0.773}            & \textbf{0.773}                                                                                                \\ \hline
\multicolumn{1}{l}{\multirow{3}{*}{\textbf{Wikipedia}}} & \multicolumn{1}{c}{\textit{10\%}}              & \textit{75\%}                                  & \underline{0.952}                      & 0.551                               & 0.949           & \textbf{0.954}                                                                                                \\ 
\multicolumn{1}{l}{}                                    & \multicolumn{1}{c}{\textit{75\%}}            & \textit{10\%}                                    & \underline{0.914}                      & 0.473                               & 0.905               & \textbf{0.927}                                                                                                \\ 
\multicolumn{1}{l}{}                                    & \multicolumn{1}{c}{\textit{75\%}}            & \textit{75\%}                                  & 0.9                                 & 0.479                               & \underline{0.906}        & \textbf{0.916}                                                                                                \\ \hline
\end{tabular}
}
\end{table}

As evident from above figures, the proposed approach (Fig.~\ref{fig:diagram}) consistently outperforms almost all other models. For example, our model achieves the best results in transductive edge prediction (Table~\ref{tab:t_e_auc}) except one case, where DyRep achieved better results. Moreover, our method achieves the best results on all datasets for inductive edge prediction (Table~\ref{tab:i_e_auc}). It seems that capturing network-wide common patterns with CAW-based features generally offers an improvement over existing models, although in some cases information added by these features may be duplicated by other encoding modules.

\begin{table}[H]
\centering
\caption{Transductive node classification, AUC-ROC; best in bold, second-best underlined}
\label{tab:t_n_auc}
\resizebox{\columnwidth}{!}{
\begin{tabular}{lccccccc}\hline
& \multicolumn{1}{c}{\textit{\textbf{Node mask}}} & \multicolumn{1}{c}{\textit{\textbf{Edge mask}}} & \multicolumn{1}{c}{\textit{\textbf{DyRep}}} & \multicolumn{1}{c}{\textit{\textbf{Jodie}}} & \multicolumn{1}{c}{\textit{\textbf{TGN}}} & \multicolumn{1}{c}{\textit{\textbf{\begin{tabular}[c]{@{}c@{}}Our \end{tabular}}}} \\ \hline
\multicolumn{1}{l}{\multirow{3}{*}{\textbf{Reddit}}}    & \multicolumn{1}{c}{\textit{10\%}}              & \textit{75\%}                                  & 0.531                                        & 0.421                                        & \underline{0.635}                                                                                                                                     & \textbf{0.659}                                                                                                         \\ 
\multicolumn{1}{l}{}                                    & \multicolumn{1}{c}{\textit{75\%}}            & \textit{10\%}                                    & \underline{0.601}                                        & 0.435                                        & 0.584                                                                                                                                      & \textbf{0.627}                                                                                                         \\ 
\multicolumn{1}{l}{}                                    & \multicolumn{1}{c}{\textit{75\%}}            & \textit{75\%}                                  & 0.597                                        & 0.439                                        & \underline{0.602}                                                                                                                                      & \textbf{0.658}                                                                                                         \\ \hline
\multicolumn{1}{l}{\multirow{3}{*}{\textbf{Wikipedia}}} & \multicolumn{1}{c}{\textit{10\%}}              & \textit{75\%}                                  & 0.771                                        & \textbf{0.842}                               & 0.735                                                                                                                                                & \underline{0.81}                                                                                                          \\ 
\multicolumn{1}{l}{}                                    & \multicolumn{1}{c}{\textit{75\%}}            & \textit{10\%}                                    & 0.769                                        & \textbf{0.836}                               & 0.738                                                                                                                                             & \underline{0.8}                                                                                                           \\ 
\multicolumn{1}{l}{}                                    & \multicolumn{1}{c}{\textit{75\%}}            & \textit{75\%}                                  & 0.755                                        & \textbf{0.855}                               & 0.8                                                                                                                                               & \underline{0.823}                                                \\ \hline
\end{tabular}
}
\end{table}

\begin{table}[H]
\centering\caption{Inductive node classification, AUC-ROC; best in bold, second-best underlined}
\label{tab:i_n_auc}
\resizebox{\columnwidth}{!}{
\begin{tabular}{lccccccc}\hline
\textbf{}                                                & \multicolumn{1}{c}{\textit{\textbf{Node mask}}} & \multicolumn{1}{c}{\textit{\textbf{Edge mask}}} & \multicolumn{1}{c}{\textit{\textbf{DyRep}}} & \multicolumn{1}{c}{\textit{\textbf{Jodie}}} & \multicolumn{1}{c}{\textit{\textbf{TGN}}} & \multicolumn{1}{c}{\textit{\textbf{\begin{tabular}[c]{@{}c@{}}Our\end{tabular}}}} \\ \hline
\multicolumn{1}{l}{\multirow{3}{*}{\textbf{Reddit}}}    & \multicolumn{1}{c}{\textit{10\%}}              & \textit{75\%}                                  & 0.51                                         & 0.456                                        & \underline{0.576}                                                                                                                                      & \textbf{0.625}                                                                                                         \\ 
\multicolumn{1}{l}{}                                    & \multicolumn{1}{c}{\textit{75\%}}            & \textit{10\%}                                    & \underline{0.544}                               & 0.539                                        & 0.495                                                                                                                                              & \textbf{0.563}                                                                                                         \\ 
\multicolumn{1}{l}{}                                    & \multicolumn{1}{c}{\textit{75\%}}            & \textit{75\%}                                  & 0.521                                        & 0.567                                        & \textbf{0.592}                                                                                                                                     & \underline{0.584}                                                                                                         \\ \hline
\multicolumn{1}{l}{\multirow{3}{*}{\textbf{Wikipedia}}} & \multicolumn{1}{c}{\textit{10\%}}              & \textit{75\%}                                  & 0.749                                        & \textbf{0.883}                               & 0.638                                                                                                                                               & \underline{0.812}                                                                                                         \\ 
\multicolumn{1}{l}{}                                    & \multicolumn{1}{c}{\textit{75\%}}            & \textit{10\%}                                    & 0.619                                        & \underline{0.73}                                & 0.649                                                                                                                                               & \textbf{0.766}                                                                                                         \\ 
\multicolumn{1}{l}{}                                    & \multicolumn{1}{c}{\textit{75\%}}            & \textit{75\%}                                  & 0.665                                        & \underline{0.743}                               & 0.597                                                                                                                                              & \textbf{0.745}                                                                                                         \\ \hline
\end{tabular}
}
\end{table}

Table \ref{tab:t_n_auc} and Table \ref{tab:i_n_auc} present the results for node classification tasks in transductive and inductive settings respectively. Though our model generally outperforms existing methods, there is one exception. Indeed, Jodie~\cite{jodie} shows the best node classification quality on Wikipedia dataset. As a matter of fact, the Jodie model incorporates both static and dynamic node embeddings, which might contributed to better encoding of interests and behavior patterns of the Wikipedia editors

\section{Banking results}

As an industrial application of the proposed framework, we chose a pre-existing problem posited by a major European banks with a corresponding dataset which consists of transactions between different companies. Each transaction has a timestamp so we can treat it as a temporal edge. This allows us to consider only "edge events". We use the following edge features: transaction amount, timestamp and type of transaction, (e.g. credit or state duty). We one-hot encode the type of transaction, which gives us 50 edge features in total: 49 from one-hot encoding and 1 from transaction amount. These features were normalized.

    \begin{figure}
        \centering
        \includegraphics[width=.8\columnwidth]{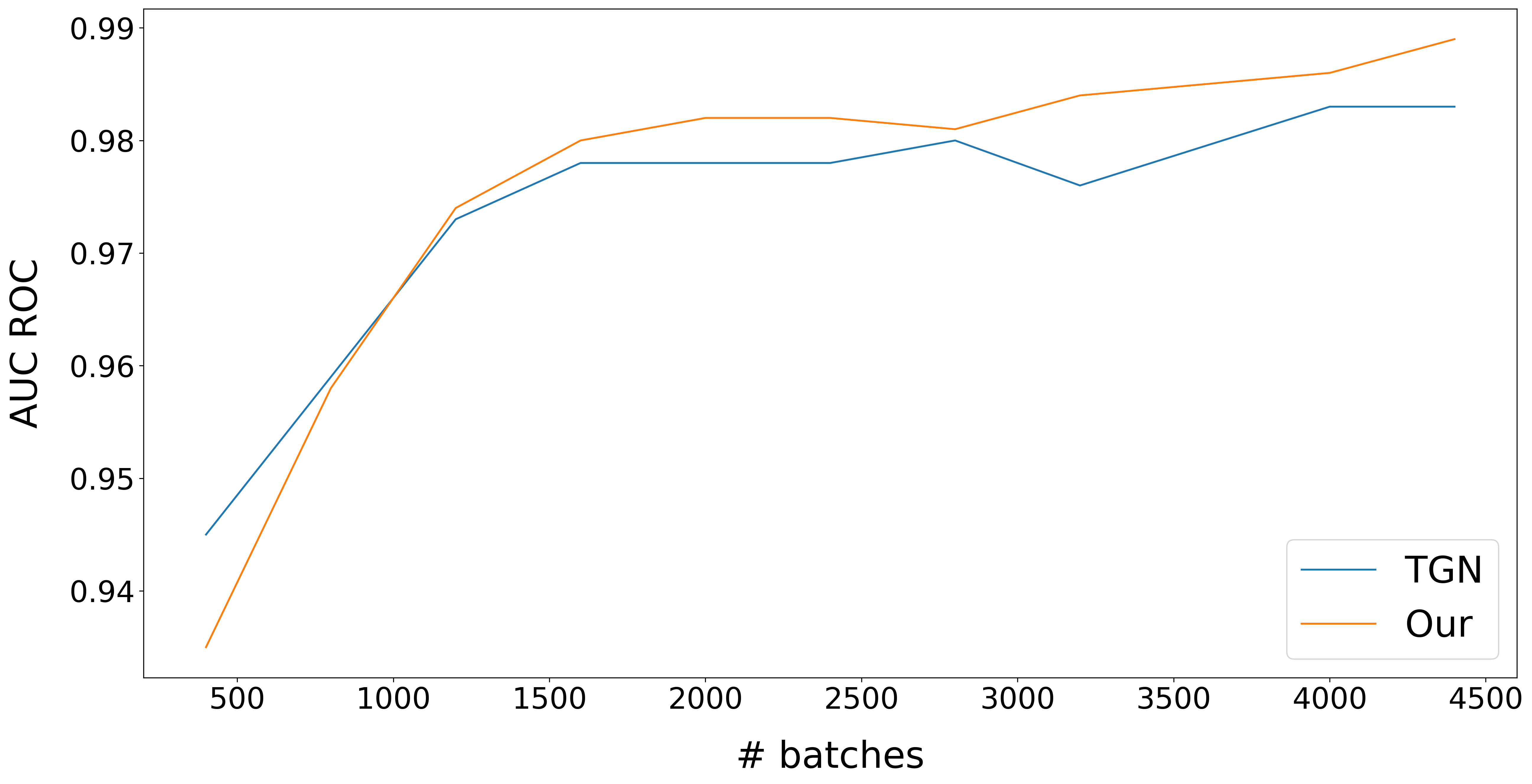}
        \caption{Inductive edge prediction on transactions of regional subdivision of major European bank}
        \label{fig:learn_curve}
    \end{figure}

At first, we compared our model with the original TGN~\cite{tgn} on a link prediction task. We obtain the training and validation datasets consisting of transactions from a regional subdivision of a major European bank for a time period about one week. This dataset contain about 2 millions of transactions. We select about 40000 newest transactions to test dataset and all the rest is given to the train i. e. test dataset transactions are always newer then train transactions. We use the following procedure to check performance of our model on test dataset. First, we split training data into small chunks (400 batches) and gradually increase the number of chunks used to train the model. Each batch contains 512 transactions. The results are presented in Fig.~\ref{fig:learn_curve}.

We can see very high performance of dynamic graph approach models. 
It can be noticed that performance of our model grows a little bit slower than for TGN~\cite{tgn}, but after about 600000 transactions our model takes the lead. This slight time lag in training is due to our model taking additional time to learn extra information about interactions between nodes via learning CAW~\cite{caw} part of our model. This extra information allows to gain better results in the end of training.

\begin{table}[!ht]
\centering
\caption{Node classification results based on node embeddings obtained from TGN and our proposed model with feature dimension $d$ being multiple of $8\times2$ shape}
\label{tab:emb_valid}
\begin{tabular}{lc}\hline
Node Embeddings &  AUC-ROC\\ \hline
TGN & $0.626 \pm 0.072$ \\
Our, $d=10$  & $0.653 \pm 0.049$ \\
Our, $d=20$  & $0.598 \pm 0.067$ \\
Our, $d=100$ & $0.578 \pm 0.094$ \\
\hline
\end{tabular}
\end{table}


The architecture of our model and TGN~\cite{tgn} model allows us to produce node embeddings which can the be used as input for a wide range of possible downstream tasks. For our experiment we have taken prediction of company default as the downstream task. The dataset contains about 5000 companies, some of which will go bankrupt in a specific period of time in the future (180 days) and others will not. To train a classification model we used LightGBM~\cite{ke2017lightgbm}. The obtained results were averaged over with 5-fold cross-validation and presented in Table \ref{tab:emb_valid}.

Here our best model has $2.7\%$ higher AUC ROC when compared to the TGN~\cite{tgn}. That means that extra information about interactions between nodes captured by CAW~\cite{caw} part of our model is useful for classification problems. We see also that best performance is observed with small CAW messages shape. This is related to the structure of our dataset. It's number of edges is comparable with number of nodes, thus the graph is sparse. Bigger CAW messages shapes can be useful for more dense graphs.

\section{Conclusion}
In this work, we proposed a novel model for temporal graph embedding (Fig.~\ref{fig:diagram}) that shows improvement over existing methods~\cite{tgn,dyrep,jodie} on various prediction tasks while preserving the ability to generate node embeddings. Moreover, we implemented a novel experimental framework (Fig.~\ref{fig:pipeline}) that can process most kinds of graph data and an arbitrary dynamic graph inference model. 
Experimental study demonstrates the applicability of our method to solving various node/edge prediction tasks on temporal networks, and significantly improving the existing results. 

In future, it is possible to improve the performance of our framework for its application to the real-life temporal graph of bank's transactions with tens of billions of nodes. It is necessary to study efficient node/edge sampling strategies, choosing those that overcome the limitations of current models when scaling to large graphs while preserving highly-informative edge features propagation. 

\section*{Acknowledgement} This research was supported in part through computational resources of HPC facilities at
NRU HSE. We thank all the colleagues and students who helps with the discussion of our study, in particular, Andrey Plyuschevskiy and Sergey Klyahandler.


\bibliographystyle{ACM-Reference-Format}
\bibliography{arxiv_paper}


\begin{thebibliography}{65}


\ifx \showCODEN    \undefined \def \showCODEN     #1{\unskip}     \fi
\ifx \showDOI      \undefined \def \showDOI       #1{#1}\fi
\ifx \showISBNx    \undefined \def \showISBNx     #1{\unskip}     \fi
\ifx \showISBNxiii \undefined \def \showISBNxiii  #1{\unskip}     \fi
\ifx \showISSN     \undefined \def \showISSN      #1{\unskip}     \fi
\ifx \showLCCN     \undefined \def \showLCCN      #1{\unskip}     \fi
\ifx \shownote     \undefined \def \shownote      #1{#1}          \fi
\ifx \showarticletitle \undefined \def \showarticletitle #1{#1}   \fi
\ifx \showURL      \undefined \def \showURL       {\relax}        \fi
\providecommand\bibfield[2]{#2}
\providecommand\bibinfo[2]{#2}
\providecommand\natexlab[1]{#1}
\providecommand\showeprint[2][]{arXiv:#2}

\bibitem[\protect\citeauthoryear{A. and J.}{A. and J.}{2016}]%
        {node2vec}
\bibfield{author}{\bibinfo{person}{Grover A.} {and} \bibinfo{person}{Leskovec
  J.}} \bibinfo{year}{2016}\natexlab{}.
\newblock \bibinfo{title}{node2vec: Scalable Feature Learning for Networks}.
\newblock
\newblock
\showeprint[arxiv]{1607.00653}~[cs.SI]


\bibitem[\protect\citeauthoryear{A. and et~al.}{A. and et~al.}{2019}]%
        {dyngan}
\bibfield{author}{\bibinfo{person}{Maheshwari A.} {and} \bibinfo{person}{et
  al.}} \bibinfo{year}{2019}\natexlab{}.
\newblock \showarticletitle{DynGAN: Generative Adversarial Networks for Dynamic
  Network Embedding}. In \bibinfo{booktitle}{\emph{Graph Representation
  Learning Workshop at NeurIPS}}.
\newblock


\bibitem[\protect\citeauthoryear{A. and et~al.}{A. and et~al.}{2020a}]%
        {egcn}
\bibfield{author}{\bibinfo{person}{Pareja A.} {and} \bibinfo{person}{et al.}}
  \bibinfo{year}{2020}\natexlab{a}.
\newblock \showarticletitle{{EvolveGCN}: Evolving Graph Convolutional Networks
  for Dynamic Graphs}.
\newblock \bibinfo{journal}{\emph{Proceedings of AAAI-20 Conference on AI}}
  \bibinfo{volume}{34} (\bibinfo{year}{2020}), \bibinfo{pages}{5363--5370}.
\newblock


\bibitem[\protect\citeauthoryear{A. and et~al.}{A. and et~al.}{2020b}]%
        {dysat}
\bibfield{author}{\bibinfo{person}{Sankar A.} {and} \bibinfo{person}{et al.}}
  \bibinfo{year}{2020}\natexlab{b}.
\newblock \showarticletitle{DySAT: Deep Neural Representation Learning on
  Dynamic Graphs via Self-Attention Networks}. In
  \bibinfo{booktitle}{\emph{Proceedings of ACM IC on WSDM}}
  \emph{(\bibinfo{series}{WSDM '20})}. \bibinfo{pages}{519–527}.
\newblock


\bibitem[\protect\citeauthoryear{B., C., and W.}{B. et~al\mbox{.}}{2019}]%
        {knyazev2019learning}
\bibfield{author}{\bibinfo{person}{Knyazev B.}, \bibinfo{person}{Augusta C.},
  {and} \bibinfo{person}{Taylor~G. W.}} \bibinfo{year}{2019}\natexlab{}.
\newblock \bibinfo{title}{Learning Temporal Attention in Dynamic Graphs with
  Bilinear Interactions}.
\newblock
\newblock
\showeprint[arxiv]{1909.10367}~[stat.ML]


\bibitem[\protect\citeauthoryear{B., R., and S.}{B. et~al\mbox{.}}{2014}]%
        {deepwalk}
\bibfield{author}{\bibinfo{person}{Perozzi B.}, \bibinfo{person}{Al-Rfou R.},
  {and} \bibinfo{person}{Skiena S.}} \bibinfo{year}{2014}\natexlab{}.
\newblock \showarticletitle{Deepwalk: Online learning of social
  representations}. In \bibinfo{booktitle}{\emph{Proceedings of ACM SIGKDD IC
  on KDD'20}}. \bibinfo{pages}{701--710}.
\newblock


\bibitem[\protect\citeauthoryear{B., V., and C.}{B. et~al\mbox{.}}{2000}]%
        {tenenbaum2000global}
\bibfield{author}{\bibinfo{person}{Tenenbaum~J. B.}, \bibinfo{person}{De~Silva
  V.}, {and} \bibinfo{person}{Langford~J. C.}} \bibinfo{year}{2000}\natexlab{}.
\newblock \showarticletitle{A global geometric framework for nonlinear
  dimensionality reduction}.
\newblock \bibinfo{journal}{\emph{science}} \bibinfo{volume}{290},
  \bibinfo{number}{5500} (\bibinfo{year}{2000}), \bibinfo{pages}{2319--2323}.
\newblock


\bibitem[\protect\citeauthoryear{B., H., and Z.}{B. et~al\mbox{.}}{2018}]%
        {ijcai2018-505}
\bibfield{author}{\bibinfo{person}{Yu B.}, \bibinfo{person}{Yin H.}, {and}
  \bibinfo{person}{Zhu Z.}} \bibinfo{year}{2018}\natexlab{}.
\newblock \showarticletitle{Spatio-Temporal Graph Convolutional Networks: A
  Deep Learning Framework for Traffic Forecasting}. In
  \bibinfo{booktitle}{\emph{Proceedings of IJCAI-18}}.
  \bibinfo{publisher}{International Joint Conferences on Artificial
  Intelligence Organization}, \bibinfo{pages}{3634--3640}.
\newblock


\bibitem[\protect\citeauthoryear{C. and et~al.}{C. and et~al.}{2019}]%
        {resgnn}
\bibfield{author}{\bibinfo{person}{Chen C.} {and} \bibinfo{person}{et al.}}
  \bibinfo{year}{2019}\natexlab{}.
\newblock \showarticletitle{Gated Residual Recurrent Graph Neural Networks for
  Traffic Prediction}.
\newblock \bibinfo{journal}{\emph{Proceedings of AAAI-19 Conference on AI}}
  \bibinfo{volume}{33}, \bibinfo{number}{01} (\bibinfo{year}{2019}),
  \bibinfo{pages}{485--492}.
\newblock


\bibitem[\protect\citeauthoryear{D. and et~al.}{D. and et~al.}{2020}]%
        {tgat}
\bibfield{author}{\bibinfo{person}{Xu D.} {and} \bibinfo{person}{et al.}}
  \bibinfo{year}{2020}\natexlab{}.
\newblock \bibinfo{title}{Inductive Representation Learning on Temporal
  Graphs}.
\newblock
\newblock
\showeprint[arxiv]{2002.07962}~[cs.LG]


\bibitem[\protect\citeauthoryear{DT, RF, Vieira, and A.}{DT
  et~al\mbox{.}}{2021}]%
        {barros2021survey}
\bibfield{author}{\bibinfo{person}{Barros~C. DT},
  \bibinfo{person}{Mendon{\c{c}}a~M. RF}, \bibinfo{person}{Alex~B Vieira},
  {and} \bibinfo{person}{Ziviani A.}} \bibinfo{year}{2021}\natexlab{}.
\newblock \bibinfo{title}{A Survey on Embedding Dynamic Graphs}.
\newblock
\newblock
\showeprint[arxiv]{2101.01229}~[cs.LG]


\bibitem[\protect\citeauthoryear{E. and et~al.}{E. and et~al.}{2020}]%
        {tgn}
\bibfield{author}{\bibinfo{person}{Rossi E.} {and} \bibinfo{person}{et al.}}
  \bibinfo{year}{2020}\natexlab{}.
\newblock \bibinfo{title}{Temporal Graph Networks for Deep Learning on Dynamic
  Graphs}.
\newblock
\newblock
\showeprint[arxiv]{2006.10637}~[cs.LG]


\bibitem[\protect\citeauthoryear{F., A., and M.}{F. et~al\mbox{.}}{2020}]%
        {cd-gcn}
\bibfield{author}{\bibinfo{person}{Manessi F.}, \bibinfo{person}{Rozza A.},
  {and} \bibinfo{person}{Manzo M.}} \bibinfo{year}{2020}\natexlab{}.
\newblock \showarticletitle{Dynamic graph convolutional networks}.
\newblock \bibinfo{journal}{\emph{Pattern Recognition}}  \bibinfo{volume}{97}
  (\bibinfo{year}{2020}), \bibinfo{pages}{107000}.
\newblock


\bibitem[\protect\citeauthoryear{Fathy and et~al.}{Fathy and et~al.}{2020}]%
        {temporalgat}
\bibfield{author}{\bibinfo{person}{A Fathy} {and} \bibinfo{person}{et al.}}
  \bibinfo{year}{2020}\natexlab{}.
\newblock \showarticletitle{TemporalGAT: Attention-Based Dynamic Graph
  Representation Learning}. In \bibinfo{booktitle}{\emph{Advances in Knowledge
  Discovery and Data Mining}}. \bibinfo{publisher}{Springer},
  \bibinfo{address}{Cham}, \bibinfo{pages}{413--423}.
\newblock
\showISBNx{978-3-030-47426-3}


\bibitem[\protect\citeauthoryear{Foundation}{Foundation}{2010}]%
        {dumpsWikimedia}
\bibfield{author}{\bibinfo{person}{Wikimedia Foundation}.}
  \bibinfo{year}{2010}\natexlab{}.
\newblock \bibinfo{title}{Wikimedia Downloads}.
\newblock \bibinfo{howpublished}{\url{http://dumps.wikimedia.org/}}.
\newblock


\bibitem[\protect\citeauthoryear{H. and et~al.}{H. and et~al.}{2021}]%
        {chen2021hili}
\bibfield{author}{\bibinfo{person}{Chen H.} {and} \bibinfo{person}{et al.}}
  \bibinfo{year}{2021}\natexlab{}.
\newblock \showarticletitle{Highly Liquid Temporal Interaction Graph
  Embeddings}. In \bibinfo{booktitle}{\emph{Proceedings of the Web Conference
  2021}}. \bibinfo{pages}{1639--1648}.
\newblock


\bibitem[\protect\citeauthoryear{H. and et~al.}{H. and et~al.}{2018}]%
        {ctdne}
\bibfield{author}{\bibinfo{person}{Nguyen~G. H.} {and} \bibinfo{person}{et
  al.}} \bibinfo{year}{2018}\natexlab{}.
\newblock \showarticletitle{Continuous-time dynamic network embeddings}. In
  \bibinfo{booktitle}{\emph{Companion Proceedings of the The Web Conference
  2018}}. \bibinfo{pages}{969--976}.
\newblock


\bibitem[\protect\citeauthoryear{H. and et~al.}{H. and et~al.}{2020}]%
        {dyhatr}
\bibfield{author}{\bibinfo{person}{Xue H.} {and} \bibinfo{person}{et al.}}
  \bibinfo{year}{2020}\natexlab{}.
\newblock \bibinfo{title}{Modeling dynamic heterogeneous network for link
  prediction using hierarchical attention with temporal rnn}.
\newblock
\newblock
\showeprint[arxiv]{2004.01024}~[cs.SI]


\bibitem[\protect\citeauthoryear{I. and et~al.}{I. and et~al.}{2021}]%
        {makarov2021survey}
\bibfield{author}{\bibinfo{person}{Makarov I.} {and} \bibinfo{person}{et al.}}
  \bibinfo{year}{2021}\natexlab{}.
\newblock \showarticletitle{Survey on graph embeddings and their applications
  to machine learning problems on graphs}.
\newblock \bibinfo{journal}{\emph{PeerJ Computer Science}}  \bibinfo{volume}{7}
  (\bibinfo{year}{2021}), \bibinfo{pages}{e357}.
\newblock


\bibitem[\protect\citeauthoryear{Ivanov and Burnaev}{Ivanov and
  Burnaev}{2018}]%
        {ivanov2018anonymous}
\bibfield{author}{\bibinfo{person}{Sergey Ivanov} {and} \bibinfo{person}{Evgeny
  Burnaev}.} \bibinfo{year}{2018}\natexlab{}.
\newblock \showarticletitle{Anonymous walk embeddings}. In
  \bibinfo{booktitle}{\emph{International Conference on Machine Learning}}.
  PMLR, \bibinfo{pages}{2186--2195}.
\newblock

\bibitem[\protect\citeauthoryear{J. and et~al.}{J. and et~al.}{2018a}]%
        {gclstm}
\bibfield{author}{\bibinfo{person}{Chen J.} {and} \bibinfo{person}{et al.}}
  \bibinfo{year}{2018}\natexlab{a}.
\newblock \bibinfo{title}{GC-LSTM: Graph Convolution Embedded LSTM for Dynamic
  Link Prediction}.
\newblock
\newblock
\showeprint[arxiv]{1812.04206}~[cs.SI]


\bibitem[\protect\citeauthoryear{J. and B.}{J. and B.}{2021}]%
        {Gao2021OnTE}
\bibfield{author}{\bibinfo{person}{Gao J.} {and} \bibinfo{person}{Ribeiro B.}}
  \bibinfo{year}{2021}\natexlab{}.
\newblock \bibinfo{title}{On the Equivalence Between Temporal and Static Graph
  Representations for Observational Predictions}.
\newblock
\newblock
\showeprint[arxiv]{2103.07016}~[cs.LG]


\bibitem[\protect\citeauthoryear{J. and et~al.}{J. and et~al.}{2008}]%
        {leskovec2008community}
\bibfield{author}{\bibinfo{person}{Leskovec J.} {and} \bibinfo{person}{et al.}}
  \bibinfo{year}{2008}\natexlab{}.
\newblock \bibinfo{title}{Community Structure in Large Networks: Natural
  Cluster Sizes and the Absence of Large Well-Defined Clusters}.
\newblock
\newblock
\showeprint[arxiv]{0810.1355}~[cs.DS]


\bibitem[\protect\citeauthoryear{J. and et~al.}{J. and et~al.}{2019}]%
        {lrgcn}
\bibfield{author}{\bibinfo{person}{Li J.} {and} \bibinfo{person}{et al.}}
  \bibinfo{year}{2019}\natexlab{}.
\newblock \showarticletitle{Predicting path failure in time-evolving graphs}.
  In \bibinfo{booktitle}{\emph{Proceedings of ACM SIGKDD IC on KDD'19}}.
  \bibinfo{pages}{1279--1289}.
\newblock


\bibitem[\protect\citeauthoryear{J. and et~al.}{J. and et~al.}{2018b}]%
        {you2018graphrnn}
\bibfield{author}{\bibinfo{person}{You J.} {and} \bibinfo{person}{et al.}}
  \bibinfo{year}{2018}\natexlab{b}.
\newblock \showarticletitle{Graphrnn: Generating realistic graphs with deep
  auto-regressive models}. In \bibinfo{booktitle}{\emph{ICML'18}}. PMLR,
  \bibinfo{pages}{5708--5717}.
\newblock


\bibitem[\protect\citeauthoryear{J. and et~al.}{J. and et~al.}{2020}]%
        {a3t-gcn}
\bibfield{author}{\bibinfo{person}{Zhu J.} {and} \bibinfo{person}{et al.}}
  \bibinfo{year}{2020}\natexlab{}.
\newblock \bibinfo{title}{A3t-gcn: Attention temporal graph convolutional
  network for traffic forecasting}.
\newblock
\newblock
\showeprint[arxiv]{2006.11583}~[cs.LG]


\bibitem[\protect\citeauthoryear{K. and et~al.}{K. and et~al.}{2019}]%
        {gcn-gan}
\bibfield{author}{\bibinfo{person}{Lei K.} {and} \bibinfo{person}{et al.}}
  \bibinfo{year}{2019}\natexlab{}.
\newblock \showarticletitle{Gcn-gan: A non-linear temporal link prediction
  model for weighted dynamic networks}. In \bibinfo{booktitle}{\emph{IEEE
  INFOCOM'19}}. IEEE, \bibinfo{pages}{388--396}.
\newblock


\bibitem[\protect\citeauthoryear{Kazemi S.~M.}{Kazemi S.~M.}{2020}]%
        {dyngraph_survey}
\bibfield{author}{\bibinfo{person}{et~al Kazemi S.~M.}}
  \bibinfo{year}{2020}\natexlab{}.
\newblock \showarticletitle{Representation Learning for Dynamic Graphs: A
  Survey}.
\newblock \bibinfo{journal}{\emph{J. Mach. Learn. Res.}} \bibinfo{volume}{21},
  \bibinfo{number}{70} (\bibinfo{year}{2020}), \bibinfo{pages}{1--73}.
\newblock


\bibitem[\protect\citeauthoryear{Ke, Meng, Finley, Wang, Chen, Ma, Ye, and
  Liu}{Ke et~al\mbox{.}}{2017}]%
        {ke2017lightgbm}
\bibfield{author}{\bibinfo{person}{Guolin Ke}, \bibinfo{person}{Qi Meng},
  \bibinfo{person}{Thomas Finley}, \bibinfo{person}{Taifeng Wang},
  \bibinfo{person}{Wei Chen}, \bibinfo{person}{Weidong Ma},
  \bibinfo{person}{Qiwei Ye}, {and} \bibinfo{person}{Tie-Yan Liu}.}
  \bibinfo{year}{2017}\natexlab{}.
\newblock \showarticletitle{Lightgbm: A highly efficient gradient boosting
  decision tree}.
\newblock \bibinfo{journal}{\emph{Advances in neural information processing
  systems}}  \bibinfo{volume}{30} (\bibinfo{year}{2017}),
  \bibinfo{pages}{3146--3154}.
\newblock


\bibitem[\protect\citeauthoryear{Kruskal}{Kruskal}{1978}]%
        {kruskal1978multidimensional}
\bibfield{author}{\bibinfo{person}{Joseph~B Kruskal}.}
  \bibinfo{year}{1978}\natexlab{}.
\newblock \bibinfo{booktitle}{\emph{Multidimensional scaling}}.
\newblock \bibinfo{publisher}{Sage}.
\newblock


\bibitem[\protect\citeauthoryear{L. and et~al.}{L. and et~al.}{2018}]%
        {dne}
\bibfield{author}{\bibinfo{person}{Du L.} {and} \bibinfo{person}{et al.}}
  \bibinfo{year}{2018}\natexlab{}.
\newblock \showarticletitle{Dynamic Network Embedding: An Extended Approach for
  Skip-gram based Network Embedding.}. In \bibinfo{booktitle}{\emph{Proceedings
  of IJCAI-18}}. \bibinfo{pages}{2086--2092}.
\newblock


\bibitem[\protect\citeauthoryear{L., R., and J.}{L. et~al\mbox{.}}{2017}]%
        {graphsage}
\bibfield{author}{\bibinfo{person}{Hamilton~W. L.}, \bibinfo{person}{Ying R.},
  {and} \bibinfo{person}{Leskovec J.}} \bibinfo{year}{2017}\natexlab{}.
\newblock \showarticletitle{Inductive representation learning on large graphs}.
  In \bibinfo{booktitle}{\emph{Proceedings of IC on NIPS'17}}.
  \bibinfo{pages}{1025--1035}.
\newblock


\bibitem[\protect\citeauthoryear{L. and et~al.}{L. and et~al.}{2020}]%
        {tdgnn}
\bibfield{author}{\bibinfo{person}{Qu L.} {and} \bibinfo{person}{et al.}}
  \bibinfo{year}{2020}\natexlab{}.
\newblock \showarticletitle{Continuous-Time Link Prediction via Temporal
  Dependent Graph Neural Network}. In \bibinfo{booktitle}{\emph{Proceedings of
  The Web Conference 2020}} (Taipei, Taiwan) \emph{(\bibinfo{series}{WWW
  '20})}. \bibinfo{publisher}{Association for Computing Machinery},
  \bibinfo{address}{NY}, \bibinfo{pages}{3026–3032}.
\newblock


\bibitem[\protect\citeauthoryear{L. and et~al.}{L. and et~al.}{2019}]%
        {t-gcn}
\bibfield{author}{\bibinfo{person}{Zhao L.} {and} \bibinfo{person}{et al.}}
  \bibinfo{year}{2019}\natexlab{}.
\newblock \showarticletitle{T-gcn: A temporal graph convolutional network for
  traffic prediction}.
\newblock \bibinfo{journal}{\emph{IEEE ITSS}} \bibinfo{volume}{21},
  \bibinfo{number}{9} (\bibinfo{year}{2019}), \bibinfo{pages}{3848--3858}.
\newblock


\bibitem[\protect\citeauthoryear{M., J., and J.}{M. et~al\mbox{.}}{2018}]%
        {ding2018semi}
\bibfield{author}{\bibinfo{person}{Ding M.}, \bibinfo{person}{Tang J.}, {and}
  \bibinfo{person}{Zhang J.}} \bibinfo{year}{2018}\natexlab{}.
\newblock \showarticletitle{Semi-supervised learning on graphs with generative
  adversarial nets}. In \bibinfo{booktitle}{\emph{Proceedings of ACM IC on
  CIKM'18}}. \bibinfo{pages}{913--922}.
\newblock


\bibitem[\protect\citeauthoryear{M. and et~al.}{M. and et~al.}{2020}]%
        {tempnode2vec}
\bibfield{author}{\bibinfo{person}{Haddad M.} {and} \bibinfo{person}{et al.}}
  \bibinfo{year}{2020}\natexlab{}.
\newblock \showarticletitle{TemporalNode2vec: Temporal Node Embedding in
  Temporal Networks}. In \bibinfo{booktitle}{\emph{Complex Networks and Their
  Applications VIII}}. \bibinfo{publisher}{Springer}, \bibinfo{address}{Cham},
  \bibinfo{pages}{891--902}.
\newblock
\showISBNx{978-3-030-36687-2}


\bibitem[\protect\citeauthoryear{M., C., and D.}{M. et~al\mbox{.}}{2008}]%
        {bipartite_stats}
\bibfield{author}{\bibinfo{person}{Latapy M.}, \bibinfo{person}{Magnien C.},
  {and} \bibinfo{person}{Vecchio~N. D.}} \bibinfo{year}{2008}\natexlab{}.
\newblock \showarticletitle{Basic notions for the analysis of large two-mode
  networks}.
\newblock \bibinfo{journal}{\emph{Social Networks}} \bibinfo{volume}{30},
  \bibinfo{number}{1} (\bibinfo{year}{2008}), \bibinfo{pages}{31--48}.
\newblock


\bibitem[\protect\citeauthoryear{M. and C.}{M. and C.}{2001}]%
        {martinez2001pca}
\bibfield{author}{\bibinfo{person}{Martinez~A. M.} {and}
  \bibinfo{person}{Kak~A. C.}} \bibinfo{year}{2001}\natexlab{}.
\newblock \showarticletitle{Pca versus lda}.
\newblock \bibinfo{journal}{\emph{IEEE PAMI}} \bibinfo{volume}{23},
  \bibinfo{number}{2} (\bibinfo{year}{2001}), \bibinfo{pages}{228--233}.
\newblock


\bibitem[\protect\citeauthoryear{N. and M.}{N. and M.}{2017}]%
        {gcn}
\bibfield{author}{\bibinfo{person}{Kipf~T. N.} {and} \bibinfo{person}{Welling
  M.}} \bibinfo{year}{2017}\natexlab{}.
\newblock \bibinfo{title}{Semi-Supervised Classification with Graph
  Convolutional Networks}.
\newblock
\newblock
\showeprint[arxiv]{1609.02907}~[cs.LG]


\bibitem[\protect\citeauthoryear{P. and G.}{P. and G.}{2008}]%
        {cortez2008using}
\bibfield{author}{\bibinfo{person}{Cortez P.} {and} \bibinfo{person}{Silva
  A.~M. G.}} \bibinfo{year}{2008}\natexlab{}.
\newblock \showarticletitle{Using data mining to predict secondary school
  student performance}.
\newblock \bibinfo{journal}{\emph{EUROSIS}} (\bibinfo{year}{2008}).
\newblock


\bibitem[\protect\citeauthoryear{P. and et~al.}{P. and et~al.}{2018}]%
        {dyngem18}
\bibfield{author}{\bibinfo{person}{Goyal P.} {and} \bibinfo{person}{et al.}}
  \bibinfo{year}{2018}\natexlab{}.
\newblock \bibinfo{title}{DynGEM: Deep Embedding Method for Dynamic Graphs}.
\newblock
\newblock
\showeprint[arxiv]{1805.11273}~[cs.SI]


\bibitem[\protect\citeauthoryear{P., G., A., A., P., and Y.}{P.
  et~al\mbox{.}}{2017}]%
        {gat}
\bibfield{author}{\bibinfo{person}{Veli{\v{c}}kovi{\'c} P.},
  \bibinfo{person}{Cucurull G.}, \bibinfo{person}{Casanova A.},
  \bibinfo{person}{Romero A.}, \bibinfo{person}{Lio P.}, {and}
  \bibinfo{person}{Bengio Y.}} \bibinfo{year}{2017}\natexlab{}.
\newblock \bibinfo{title}{Graph attention networks}.
\newblock
\newblock
\showeprint[arxiv]{1710.10903}~[stat.ML]


\bibitem[\protect\citeauthoryear{R. and et~al.}{R. and et~al.}{2019a}]%
        {goel2019}
\bibfield{author}{\bibinfo{person}{Goel R.} {and} \bibinfo{person}{et al.}}
  \bibinfo{year}{2019}\natexlab{a}.
\newblock \bibinfo{title}{Diachronic Embedding for Temporal Knowledge Graph
  Completion}.
\newblock
\newblock
\showeprint[arxiv]{1907.03143}~[cs.LG]


\bibitem[\protect\citeauthoryear{R.}{R.}{2018}]%
        {hisano2018semi}
\bibfield{author}{\bibinfo{person}{Hisano R.}} \bibinfo{year}{2018}\natexlab{}.
\newblock \showarticletitle{Semi-supervised graph embedding approach to dynamic
  link prediction}. In \bibinfo{booktitle}{\emph{International Workshop on
  Complex Networks}}. Springer, \bibinfo{pages}{109--121}.
\newblock


\bibitem[\protect\citeauthoryear{R. and et~al.}{R. and et~al.}{2019b}]%
        {dyrep}
\bibfield{author}{\bibinfo{person}{Trivedi R.} {and} \bibinfo{person}{et al.}}
  \bibinfo{year}{2019}\natexlab{b}.
\newblock \showarticletitle{DyRep: Learning Representations over Dynamic
  Graphs}. In \bibinfo{booktitle}{\emph{ICLR (Poster)}}.
  \bibinfo{publisher}{OpenReview.net}, \bibinfo{address}{LA}.
\newblock


\bibitem[\protect\citeauthoryear{Rosenberg}{Rosenberg}{1981}]%
        {graph_embed_issues}
\bibfield{author}{\bibinfo{person}{Arnold~L. Rosenberg}.}
  \bibinfo{year}{1981}\natexlab{}.
\newblock \showarticletitle{Issues in the study of graph embeddings}. In
  \bibinfo{booktitle}{\emph{Graphtheoretic Concepts in Computer Science}},
  \bibfield{editor}{\bibinfo{person}{Hartmut Noltemeier}} (Ed.).
  \bibinfo{publisher}{Springer}, \bibinfo{address}{Berlin},
  \bibinfo{pages}{150--176}.
\newblock
\showISBNx{978-3-540-38435-9}


\bibitem[\protect\citeauthoryear{S. and et~al.}{S. and et~al.}{2018}]%
        {to-gae}
\bibfield{author}{\bibinfo{person}{Bonner S.} {and} \bibinfo{person}{et al.}}
  \bibinfo{year}{2018}\natexlab{}.
\newblock \showarticletitle{Temporal graph offset reconstruction: Towards
  temporally robust graph representation learning}. In
  \bibinfo{booktitle}{\emph{IEEE Big Data 2018}}. IEEE,
  \bibinfo{pages}{3737--3746}.
\newblock


\bibitem[\protect\citeauthoryear{S. and et~al.}{S. and et~al.}{2019}]%
        {tna}
\bibfield{author}{\bibinfo{person}{Bonner S.} {and} \bibinfo{person}{et al.}}
  \bibinfo{year}{2019}\natexlab{}.
\newblock \showarticletitle{Temporal neighbourhood aggregation: Predicting
  future links in temporal graphs via recurrent variational graph
  convolutions}. In \bibinfo{booktitle}{\emph{2019 IEEE International
  Conference on Big Data (Big Data)}}. IEEE, \bibinfo{pages}{5336--5345}.
\newblock


\bibitem[\protect\citeauthoryear{S. and et~al.}{S. and et~al.}{1990}]%
        {deerwester1990indexing}
\bibfield{author}{\bibinfo{person}{Deerwester S.} {and} \bibinfo{person}{et
  al.}} \bibinfo{year}{1990}\natexlab{}.
\newblock \showarticletitle{Indexing by latent semantic analysis}.
\newblock \bibinfo{journal}{\emph{Journal of the ASIS\&T}}
  \bibinfo{volume}{41}, \bibinfo{number}{6} (\bibinfo{year}{1990}),
  \bibinfo{pages}{391--407}.
\newblock


\bibitem[\protect\citeauthoryear{S. and et~al.}{S. and et~al.}{2020}]%
        {ehna}
\bibfield{author}{\bibinfo{person}{Huang S.} {and} \bibinfo{person}{et al.}}
  \bibinfo{year}{2020}\natexlab{}.
\newblock \showarticletitle{Temporal network representation learning via
  historical neighborhoods aggregation}. In \bibinfo{booktitle}{\emph{IEEE
  ICDE'20}}. IEEE, \bibinfo{pages}{1117--1128}.
\newblock


\bibitem[\protect\citeauthoryear{S., X., and J.}{S. et~al\mbox{.}}{2019b}]%
        {jodie}
\bibfield{author}{\bibinfo{person}{Kumar S.}, \bibinfo{person}{Zhang X.}, {and}
  \bibinfo{person}{Leskovec J.}} \bibinfo{year}{2019}\natexlab{b}.
\newblock \bibinfo{title}{Predicting Dynamic Embedding Trajectory in Temporal
  Interaction Networks}.
\newblock
\newblock


\bibitem[\protect\citeauthoryear{S., S., and A.}{S. et~al\mbox{.}}{2019a}]%
        {dyn-vgae}
\bibfield{author}{\bibinfo{person}{Mahdavi S.}, \bibinfo{person}{Khoshraftar
  S.}, {and} \bibinfo{person}{An A.}} \bibinfo{year}{2019}\natexlab{a}.
\newblock \bibinfo{title}{Dynamic joint variational graph autoencoders}.
\newblock
\newblock
\showeprint[arxiv]{1910.01963}~[cs.LG]


\bibitem[\protect\citeauthoryear{T. and et~al.}{T. and et~al.}{2013}]%
        {word2vec}
\bibfield{author}{\bibinfo{person}{Mikolov T.} {and} \bibinfo{person}{et al.}}
  \bibinfo{year}{2013}\natexlab{}.
\newblock \bibinfo{title}{Efficient estimation of word representations in
  vector space}.
\newblock
\newblock
\showeprint[arxiv]{1301.3781}~[cs.CL]


\bibitem[\protect\citeauthoryear{T. and K.}{T. and K.}{2000}]%
        {roweis2000nonlinear}
\bibfield{author}{\bibinfo{person}{Roweis~S. T.} {and} \bibinfo{person}{Saul~L.
  K.}} \bibinfo{year}{2000}\natexlab{}.
\newblock \showarticletitle{Nonlinear dimensionality reduction by locally
  linear embedding}.
\newblock \bibinfo{journal}{\emph{science}} \bibinfo{volume}{290},
  \bibinfo{number}{5500} (\bibinfo{year}{2000}), \bibinfo{pages}{2323--2326}.
\newblock


\bibitem[\protect\citeauthoryear{U. and J.}{U. and J.}{2008}]%
        {Sharan2008}
\bibfield{author}{\bibinfo{person}{Sharan U.} {and} \bibinfo{person}{Neville
  J.}} \bibinfo{year}{2008}\natexlab{}.
\newblock \showarticletitle{Temporal-Relational Classifiers for Prediction in
  Evolving Domains}.
\newblock \bibinfo{journal}{\emph{Eighth IEEE ICDM'08}} (\bibinfo{year}{2008}),
  \bibinfo{pages}{540--549}.
\newblock


\bibitem[\protect\citeauthoryear{X. and et~al.}{X. and et~al.}{2021}]%
        {apan}
\bibfield{author}{\bibinfo{person}{Wang X.} {and} \bibinfo{person}{et al.}}
  \bibinfo{year}{2021}\natexlab{}.
\newblock \bibinfo{title}{APAN: Asynchronous Propagation Attention Network for
  Real-time Temporal Graph Embedding}.
\newblock
\newblock
\showeprint[arxiv]{2011.11545}~[cs.AI]


\bibitem[\protect\citeauthoryear{Y. and et~al.}{Y. and et~al.}{2018}]%
        {dcrnn}
\bibfield{author}{\bibinfo{person}{Li Y.} {and} \bibinfo{person}{et al.}}
  \bibinfo{year}{2018}\natexlab{}.
\newblock \bibinfo{title}{Diffusion convolutional recurrent neural network:
  Data-driven traffic forecasting}.
\newblock
\newblock
\showeprint[arxiv]{1707.01926}~[cs.LG]


\bibitem[\protect\citeauthoryear{Y. and et~al.}{Y. and et~al.}{2019}]%
        {mdne}
\bibfield{author}{\bibinfo{person}{Lu Y.} {and} \bibinfo{person}{et al.}}
  \bibinfo{year}{2019}\natexlab{}.
\newblock \showarticletitle{Temporal Network Embedding with Micro- and
  Macro-Dynamics}. In \bibinfo{booktitle}{\emph{Proceedings of the ACM IC on
  CIKM'19}} (Beijing, China) \emph{(\bibinfo{series}{CIKM '19})}.
  \bibinfo{publisher}{Association for Computing Machinery},
  \bibinfo{address}{NY}, \bibinfo{pages}{469–478}.
\newblock


\bibitem[\protect\citeauthoryear{Y., Z., Z., J., and D.}{Y.
  et~al\mbox{.}}{2020b}]%
        {dgnn}
\bibfield{author}{\bibinfo{person}{Ma Y.}, \bibinfo{person}{Guo Z.},
  \bibinfo{person}{Ren Z.}, \bibinfo{person}{Tang J.}, {and}
  \bibinfo{person}{Yin D.}} \bibinfo{year}{2020}\natexlab{b}.
\newblock \showarticletitle{Streaming Graph Neural Networks}. In
  \bibinfo{booktitle}{\emph{Proceedings of ACM IC on SIGIR'20}} (Virtual Event,
  China) \emph{(\bibinfo{series}{SIGIR '20})}. \bibinfo{publisher}{Association
  for Computing Machinery}, \bibinfo{address}{NY}, \bibinfo{pages}{719–728}.
\newblock
\showISBNx{9781450380164}


\bibitem[\protect\citeauthoryear{Y. and et~al.}{Y. and et~al.}{2016}]%
        {gcrn}
\bibfield{author}{\bibinfo{person}{Seo Y.} {and} \bibinfo{person}{et al.}}
  \bibinfo{year}{2016}\natexlab{}.
\newblock \bibinfo{title}{Structured Sequence Modeling with Graph Convolutional
  Recurrent Networks}.
\newblock
\newblock
\showeprint[arxiv]{1612.07659}~[stat.ML]


\bibitem[\protect\citeauthoryear{Y. and et~al.}{Y. and et~al.}{2021}]%
        {caw}
\bibfield{author}{\bibinfo{person}{Wang Y.} {and} \bibinfo{person}{et al.}}
  \bibinfo{year}{2021}\natexlab{}.
\newblock \bibinfo{title}{Inductive Representation Learning in Temporal
  Networks via Causal Anonymous Walks}.
\newblock
\newblock
\showeprint[arxiv]{2101.05974}~[cs.LG]


\bibitem[\protect\citeauthoryear{Y., Y., and Y.}{Y. et~al\mbox{.}}{2020a}]%
        {ti-gcn}
\bibfield{author}{\bibinfo{person}{Xiang Y.}, \bibinfo{person}{Xiong Y.}, {and}
  \bibinfo{person}{Zhu Y.}} \bibinfo{year}{2020}\natexlab{a}.
\newblock \showarticletitle{TI-GCN: A Dynamic Network Embedding Method with
  Time Interval Information}. In \bibinfo{booktitle}{\emph{IEEE Big Data
  2020}}. IEEE, \bibinfo{pages}{838--847}.
\newblock


\bibitem[\protect\citeauthoryear{Yun, Jeong, Kim, Kang, and Kim}{Yun
  et~al\mbox{.}}{2019}]%
        {yun2019graph}
\bibfield{author}{\bibinfo{person}{Seongjun Yun}, \bibinfo{person}{Minbyul
  Jeong}, \bibinfo{person}{Raehyun Kim}, \bibinfo{person}{Jaewoo Kang}, {and}
  \bibinfo{person}{Hyunwoo~J Kim}.} \bibinfo{year}{2019}\natexlab{}.
\newblock \showarticletitle{Graph transformer networks}.
\newblock \bibinfo{journal}{\emph{Advances in Neural Information Processing
  Systems}}  \bibinfo{volume}{32} (\bibinfo{year}{2019}),
  \bibinfo{pages}{11983--11993}.
\newblock


\bibitem[\protect\citeauthoryear{Z. and et~al.}{Z. and et~al.}{2021a}]%
        {cui2021dygcn}
\bibfield{author}{\bibinfo{person}{Cui Z.} {and} \bibinfo{person}{et al.}}
  \bibinfo{year}{2021}\natexlab{a}.
\newblock \bibinfo{title}{DyGCN: Dynamic Graph Embedding with Graph
  Convolutional Network}.
\newblock
\newblock
\showeprint[arxiv]{2104.02962}~[cs.LG]


\bibitem[\protect\citeauthoryear{Z. and et~al.}{Z. and et~al.}{2021b}]%
        {tsn}
\bibfield{author}{\bibinfo{person}{Liu Z.} {and} \bibinfo{person}{et al.}}
  \bibinfo{year}{2021}\natexlab{b}.
\newblock \showarticletitle{Motif-Preserving Dynamic Attributed Network
  Embedding}. In \bibinfo{booktitle}{\emph{Proceedings of the Web Conference
  2021}} \emph{(\bibinfo{series}{WWW '21})}. \bibinfo{publisher}{Association
  for Computing Machinery}, \bibinfo{pages}{1629–1638}.
\newblock


\bibitem[\protect\citeauthoryear{Z. and et~al.}{Z. and et~al.}{2020}]%
        {tigecmn}
\bibfield{author}{\bibinfo{person}{Zhang Z.} {and} \bibinfo{person}{et al.}}
  \bibinfo{year}{2020}\natexlab{}.
\newblock \showarticletitle{Learning Temporal Interaction Graph Embedding via
  Coupled Memory Networks}. In \bibinfo{booktitle}{\emph{Proceedings of The Web
  Conference 2020}} \emph{(\bibinfo{series}{WWW '20})}.
  \bibinfo{publisher}{Association for Computing Machinery},
  \bibinfo{pages}{3049–3055}.
\newblock


\end{thebibliography}

\end{document}